\PassOptionsToPackage{table}{xcolor}
\PassOptionsToPackage{normalem}{ulem}
\documentclass{style}
\usepackage{amsmath,amssymb,mathtools}
\usepackage{float,xspace}
\usepackage{xurl}
\fancypagestyle{firststyle}{%
  \fancyhf{}%
  \fancyhead[C]{%
    \makebox[\textwidth][c]{%
      \schoollogo[0.165]{14mm}{illinois.png}\hspace{2mm}
      \raisebox{1.2mm}[0pt][0pt]{\schoollogo[0.165]{14mm}{rutgers.png}}\hspace{0.3mm}
      \schoollogo[0.206]{14mm}{gatech.png}\hfill
      \raisebox{-0.3mm}[0pt][0pt]{\schoollogo[0.19]{12mm}{logo-v5-tight.pdf}}%
    }%
  }%
}

\newcommand{\modelname}{SAEPO\xspace}
\newcommand{\tabref}[1]{Table~\ref{#1}}
\newcommand{\secref}[1]{Sec.~\ref{#1}}
\newcommand{\appref}[1]{Appendix~\ref{#1}}
\newcommand{\figref}[1]{Fig.~\ref{#1}}
\newcommand{\eqnref}[1]{Eqn. (\ref{#1})}
\renewcommand{\tilde}{\widetilde}
\renewcommand{\hat}{\widehat}
\newcommand{\green}[1]{\textcolor{green}{#1}}
\newcommand{\red}[1]{\textcolor{red}{#1}}
\newcommand{\gray}[1]{\textcolor{gray}{#1}}

\title{SAEPO: Sparse-Autoencoder-Guided\\Credit Assignment for Policy Optimization}
\hypersetup{pdftitle={SAEPO: Sparse-Autoencoder-Guided Credit Assignment for Policy Optimization},
  pdfauthor={Yusong Zhao, Zihao Zhao, Zirui Zhang, Tunyu Zhang, Ziyan Wang, Yi Wang, Chengzhi Mao, Hao Wang}}

\renewcommand{\authorlist}{%
  \authorfont\sffamily\bfseries
  \mbox{Yusong Zhao\textsuperscript{1}},
  \mbox{Zihao Zhao\textsuperscript{1}},
  \mbox{Zirui Zhang\textsuperscript{2}},
  \mbox{Tunyu Zhang\textsuperscript{2}},\\[2pt]
  \mbox{Ziyan Wang\textsuperscript{3}},
  \mbox{Yi Wang\textsuperscript{1}},
  \mbox{Chengzhi Mao\textsuperscript{2}},
  \mbox{Hao Wang\textsuperscript{1,2}}%
}

\renewcommand{\affiliationlist}{%
  \affiliationformat[1]{University of Illinois Urbana-Champaign},
  \affiliationformat[2]{Rutgers University},
  \affiliationformat[3]{Georgia Institute of Technology}%
}

\checkdata[Contact]{yusongz@illinois.edu}

\abstract{
Reinforcement Learning with Verifiable Rewards (RLVR) has demonstrated strong performance in improving large language model (LLM) reasoning, yet its credit-assignment signal remains coarse. Outcome verifiers provide scalable supervision by judging whether a completed response is correct, but they do not directly indicate which intermediate reasoning tokens contribute to success or failure. As a result, policy optimization can be inefficient, especially when training base models directly from sparse outcome feedback. In this paper, we introduce \modelname{}, a \textbf{S}parse \textbf{A}uto\textbf{E}ncoder-guided token-level credit assignment method for outcome-supervised \textbf{P}olicy \textbf{O}ptimization. Rather than treating correctness only as a rollout-level (sequence-level) signal, \modelname{} uses a pretrained sparse autoencoder (SAE) to build token-level semantic representations within reasoning trajectories; it scores each token based on the SAE concepts (e.g., penalizing wrong ``final-answer'' tokens but rewarding correct ``reasoning-process'' tokens), learned in an unsupervised manner, to achieve fine-grained credit assignment.  \modelname{} is a plug-in advantage-shaping module compatible with various policy optimization algorithms. We instantiate \modelname{} with both GRPO and GSPO, covering token-ratio and sequence-ratio policy-gradient objectives, and evaluate \modelname{} across three LLM backbones on six math and two code benchmarks.
Empirical results show that \modelname{} improves both the math and code benchmark averages over the matched baselines, with gains of up to 6.63\% and 5.57\%, respectively. On {\texttt{Qwen3-1.7B-Base}}, training curves also show faster early accuracy growth. Feature and token-level analyses suggest that the SAE-derived signal favors question-relevant concepts and reasoning tokens.
}

\begin{document}
\maketitle

\section{Introduction}

\begin{figure}[t]
  \centering

  \includegraphics[width=\linewidth]{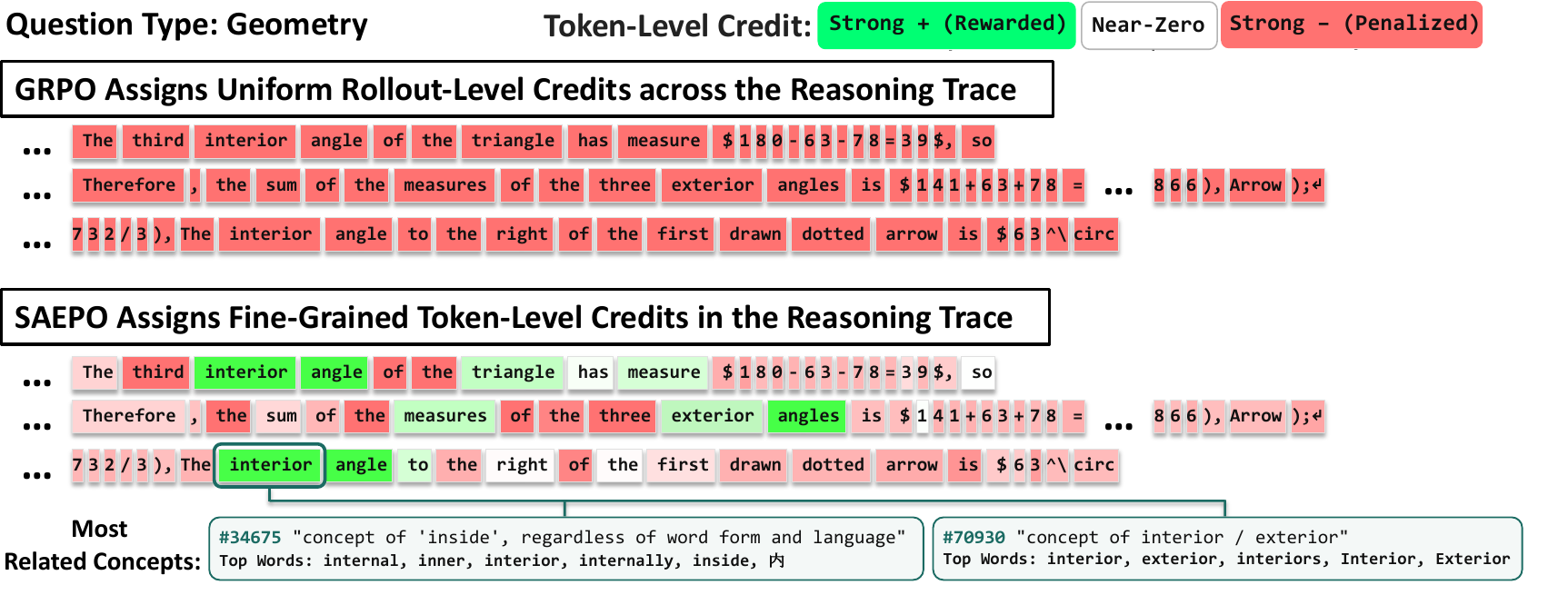}

  \caption{Comparison of credit assignment mechanisms in outcome-supervised policy optimizations. \textbf{GRPO} assigns the same rollout-level credit to all tokens in a generated reasoning trace, so decisive reasoning steps and incidental tokens receive identical credit. Our \textbf{\modelname{}} assigns question-local, token-level semantic credit based on semantic representations from the SAE feature space; therefore it rewards question-relevant reasoning tokens while penalizing uninformative or misleading tokens with more fine-grained credit assignment. {Top words of the most related concepts to an example rewarded token ``interior'' show that our SAEPO can achieve interpretable token-level credit assignment.}
}
  \label{fig:teaser}

\end{figure}

Reinforcement Learning with Verifiable Rewards (RLVR) has become an effective paradigm for improving the reasoning ability of Large Language Models (LLMs) on tasks with rule-based correctness signals. In mathematical reasoning \citep{shao2024deepseekmathpushinglimitsmathematical, Guo_2025, yu2025dapoopensourcellmreinforcement, wen2025reinforcementlearningverifiablerewards, wang2025reinforcementlearningreasoninglarge}, code generation \citep{le2022coderlmasteringcodegeneration, dai2025processsupervisionguidedpolicyoptimization, wang2025coevolvingllmcoderunit, jiang2026coderlimprovingcodegeneration}, and theorem proving \citep{lample2022hypertreeproofsearchneural, zhang2025leanabellproverposttrainingscalingformal, ji2025leanabellproverv2verifierintegratedreasoningformal}, outcome verifiers can judge whether a completed response is correct without requiring human-written reasoning traces.
This makes RLVR particularly attractive for training models on challenging reasoning tasks. Recent outcome-supervised policy-gradient methods, including GRPO \citep{shao2024deepseekmathpushinglimitsmathematical} and GSPO \citep{zheng2025gspo}, have made this recipe practical by updating the model from sampled solutions and verifiable final-answer rewards. Although these methods differ in their policy-ratio and clipping designs, their supervision ultimately arrives at the response level: a completed rollout (sequence) is judged correct or incorrect,
and the same resulting scalar learning signal must then be broadcast into distinct token-level policy-gradient updates, as shown in~\figref{fig:teaser}(top).

This broadcast is the source of the credit-assignment bottleneck. The KTAE study by \citet{sun2025ktae} observes that GRPO- and DAPO-style methods compute rollout-level advantages and assign the same value to every token in the sequence, failing to capture token-specific contributions. OAR similarly describes standard GRPO as propagating group-level rewards uniformly to every token, neglecting the varying contribution of individual reasoning steps \citep{li2026outcomegroundedadvantagereshapingfinegrained}. GTPO/GRPO-S frames the same issue as a coarse-grained assignment that applies a uniform reward to all tokens in long-chain reasoning \citep{tan2025gtpo}. These observations point to the same limitation: outcome supervision separates successful from failed rollouts, but does not explain how credit should be distributed within each rollout. %

To obtain finer-grained credit, we need a signal that can distinguish the internal reasoning states of successful and unsuccessful rollouts, rather than relying only on their final scalar outcomes. The model's hidden states provide a natural source of such information, but dense activations are difficult to compare and interpret directly. Sparse Autoencoders (SAEs) \citep{bricken2023monosemanticity,cunningham2023sparseautoencodershighlyinterpretable,templeton2024scaling,lieberum2024gemmascopeopensparse} offer a scalable way to project dense hidden states into sparse latent features, which often correspond to \emph{more interpretable} and \emph{less noisy} components of model computation. This sparse feature space gives us a semantic interface for RLVR: successful and unsuccessful rollouts of the same question can be contrasted not only by their token identities or final rewards, but by the internal features activated along their reasoning trajectories.

Based on this intuition, it is natural to use the sparse feature space to construct token-level credit signals during policy optimization. We introduce \modelname{}, a \textbf{S}parse \textbf{A}uto\textbf{E}ncoder-guided token-level credit assignment method for outcome-supervised \textbf{P}olicy \textbf{O}ptimization. Rather than treating correctness only as a rollout-level signal, \modelname{} compares correct and incorrect rollouts of the same question in semantic space constructed with SAE.
Intuitively, in each group of rollouts during GRPO variants, each token's reward depends on how many of its similar tokens are in correct and incorrect rollouts; a token with more similar tokens in correct rollouts than in incorrect rollouts receives a higher reward (more details in \figref{fig:teaser}(bottom) and a more sophisticated algorithm in \secref{sec:saepo-method}).

This design makes \modelname{} a plug-in for different policy optimization algorithms without requiring a separately trained reward model. Specifically, the SAE-derived score only reshapes the token-level advantage used during training; the verifier reward, policy ratio, clipping rule, and KL regularizer remain unchanged. As a result, the same \modelname{} module can be inserted into different outcome-supervised policy-gradient algorithms, including token-ratio methods such as GRPO \citep{shao2024deepseekmathpushinglimitsmathematical} and sequence-ratio methods such as GSPO \citep{zheng2025gspo}.
\modelname{} incurs minimal inference overhead: the SAE is used only during fine-tuning; therefore the final policy requires no SAE computation at inference time.

We instantiate SAEPO with GRPO and GSPO and evaluate it on {both mathematical reasoning and coding tasks} using \texttt{Qwen3-\allowbreak{}1.7B-\allowbreak{}Base}, \texttt{Qwen3-\allowbreak{}4B-\allowbreak{}Base} and {\texttt{OLMo-\allowbreak{}3-\allowbreak{}7B-\allowbreak{}Base}}. Across these settings, SAEPO improves {average accuracy on both math and code benchmarks} over the corresponding vanilla algorithms and {shows improved training efficiency}. Further analysis shows that the SAE-derived signal concentrates on semantically meaningful reasoning positions, suggesting that sparse internal representations provide a useful interface for fine-grained credit assignment in outcome-supervised RLVR. Our contributions are summarized as follows:

\begin{itemize}[nosep]
    \item We propose \modelname{}, a token-level credit-assignment method that uses SAE features to construct question-local contrasts between correct and incorrect rollouts.

    \item We show that \modelname{} can be used as a plug-in advantage-shaping module for different policy optimization algorithms, without modifying the verifier reward, policy ratio, clipping rule, or KL regularizer.

    \item Interpretability analyses of the SAE-derived credit signal show that \modelname{} concentrates on semantically meaningful reasoning positions and {highlights question-relevant SAE features}, offering a transparent view of how token-level credit is assigned.

    \item Experiments with different LLM backbones show improved average accuracy and {training efficiency} over the corresponding base policy optimization algorithms.
\end{itemize}

\section{Related Work}

\textbf{RLVR and Policy Optimization for Reasoning.}
Reinforcement Learning with Verifiable Rewards (RLVR) has become a central paradigm for improving LLM reasoning on tasks with rule-based correctness signals, including mathematical reasoning, code generation, and theorem proving. Recent systems such as DeepSeek-R1 \citep{Guo_2025}, Kimi k1.5 \citep{kimi2025k15}, and Qwen2.5-Math \citep{yang2024qwen25math} demonstrate the effectiveness of outcome-supervised post-training at scale. Algorithmically, GRPO \citep{shao2024deepseekmathpushinglimitsmathematical} replaces PPO's learned critic with a group-normalized baseline computed from multiple rollouts of the same question. Subsequent variants refine different aspects of this recipe: DAPO \citep{yu2025dapoopensourcellmreinforcement} introduces decoupled clipping, dynamic sampling, and token-level loss aggregation; Dr.~GRPO \citep{liu2025understandingr1zero} analyzes length and group-normalization biases; REINFORCE++ \citep{hu2025reinforceplusplus} revisits normalized policy-gradient updates without a learned critic; GSPO \citep{zheng2025gspo} moves importance sampling and clipping to the sequence level. These methods differ in their ratio construction, clipping rule, and normalization scheme, but they remain driven by outcome-level correctness signals. In contrast, our work focuses on improving the within-trajectory token-level credit assignment of such outcome-supervised updates.

\textbf{Fine-Grained Credit Assignment from Outcome Rewards.}
{A growing line of work seeks to extract denser supervision from coarse outcome feedback.
Process Reward Models (PRMs) provide step-level feedback using human-labeled or automatically generated supervision \citep{lightman2023verify,wang2024mathshepherd,zhang2024restmcts}, but require an auxiliary reward model and additional annotation or search.
Other methods infer denser training signals from preference or outcome data:
PRIME \citep{cui2025prime} trains an implicit process reward model online,
RTO \citep{zhong2024dpomeetsppo} extracts rewards from a DPO-style implicit reward model,
and offline preference-optimization methods such as cDPO \citep{lin2024critical},
Step-DPO \citep{lai2024stepdpo}, TDPO \citep{zeng2024tokenleveldpo}, and
SVPO \citep{chen2024stepvalue} modify the preference-learning objective to better account for multi-step responses.
These methods rely on auxiliary reward modeling, additional supervision, or preference-objective design. {\modelname{} instead uses a frozen SAE to shape token-level advantages while keeping the verifier reward, policy ratio, and clipping rule unchanged.}}

A second family constructs training signals directly from sampled rollouts without a separate reward model.
VinePPO \citep{kazemnejad2024vineppo} uses Monte Carlo rollouts from intermediate states,
Beyond-80/20 \citep{wang2025beyond} focuses updates on high-entropy forking positions,
GTPO \citep{tan2025gtpo} uses entropy and correctness asymmetry,
Token Hidden Reward \citep{deng2025tokenhidden} reweights advantages by likelihood dynamics,
and KTAE \citep{sun2025ktae} computes bonuses from correct-versus-incorrect token occurrence statistics.
{\modelname{} differs from these approaches in the signal it uses}: it (1) derives fine-grained signals by contrasting rollouts in a sparse semantic space, rather than over token identity or likelihood statistics, and (2) provides concept-level interpretability {through its SAE features}.

\textbf{Sparse Autoencoders for Representation-Level Reasoning and Control.}
Sparse Autoencoders (SAEs) decompose dense LLM activations into sparse latent features and have become a standard tool for mechanistic interpretability \citep{bricken2023monosemanticity,cunningham2023sparseautoencodershighlyinterpretable,templeton2024scaling,lieberum2024gemmascopeopensparse}. Architectural improvements such as Top-K SAEs \citep{gao2024scalingevaluatingsparseautoencoders}, Gated SAEs \citep{rajamanoharan2024improvingdictionarylearninggated}, and JumpReLU SAEs \citep{rajamanoharan2024jumpingaheadimprovingreconstruction} further improve reconstruction quality and feature sparsity. These developments make SAE features useful not only as post-hoc explanations, but also as structured coordinates for analyzing and manipulating model behavior. This view connects to a broader line of representation-level control, where behavioral directions are obtained by contrasting internal activations from positive and negative examples. Methods such as Contrastive Activation Addition \citep{rimsky2024steering}, Inference-Time Intervention \citep{li2023inferencetime}, and Representation Engineering \citep{zou2023representation} show that such contrasts can isolate behaviorally meaningful directions in activation space. Recent SAE-based reasoning work applies this idea in sparse feature space: ReasonScore studies SAE features associated with uncertainty, exploration, and reflection \citep{galichin2025icoveredbaseshere}; SAE-Steering controls reasoning strategies such as backtracking and cross-verification \citep{fang2026controllable}; and ReSA uses SAE-Tuning to transfer reasoning abilities without reasoning traces \citep{wang2025resatransparentreasoningmodels}. SAE features have also been incorporated into alignment and reward-modeling pipelines, including FPO \citep{yin2024directpreference}, SARM \citep{zhang2025interpretable}, SAFER \citep{shi2025safer}, and SparseRM \citep{liu2025sparsermlightweightpreferencemodeling}.
{These methods use SAE features for interpretation, steering, tuning, or reward model training.} In contrast, our \modelname{} {uses SAE to shape token-level advantages during policy optimization}.

\begin{figure}[t]
  \centering

  \includegraphics[width=\linewidth]{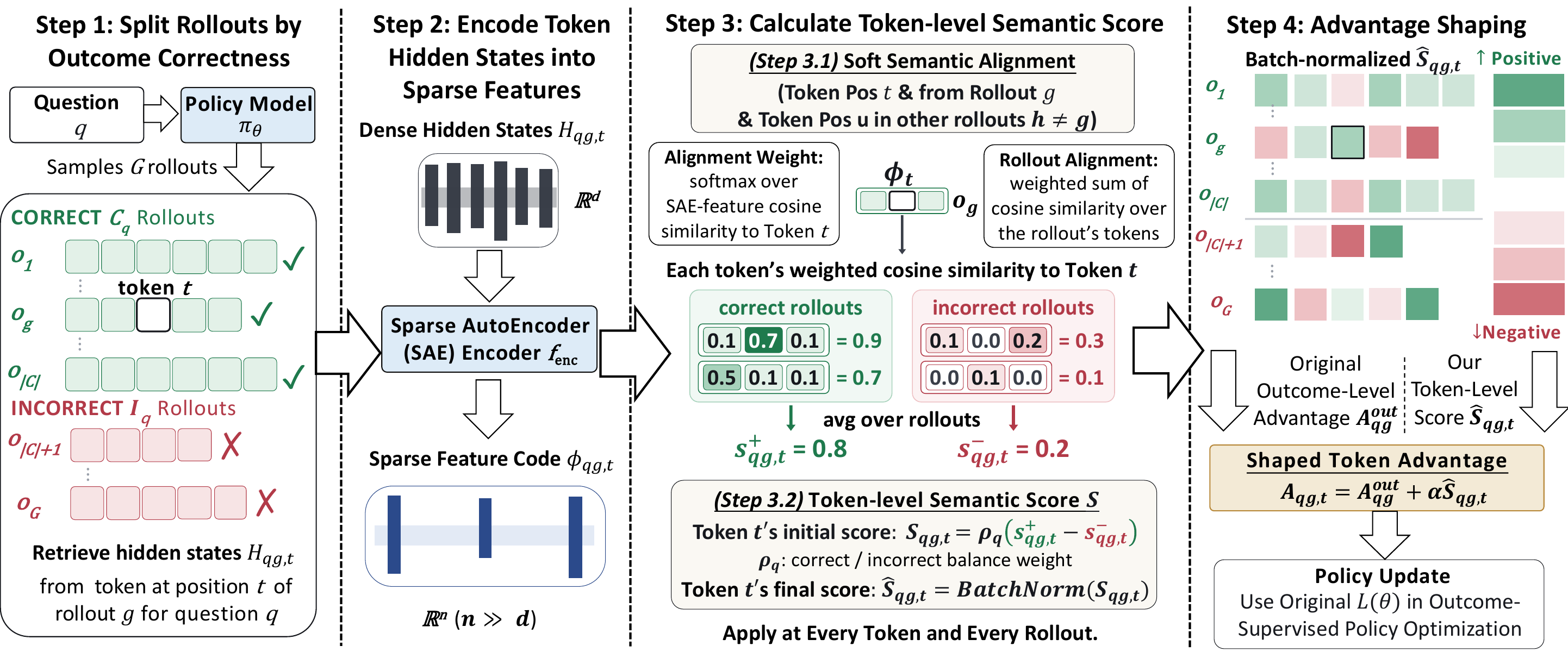}

  \caption{
    Overview of \modelname{}.
    A frozen SAE encodes token hidden states from grouped rollouts into sparse features.
    For each token, {its counts of similar tokens in correct and incorrect rollouts determine its token-level reward, with higher reward assigned when more similar tokens appear in correct rollouts than in incorrect ones}.
The {token-level reward} is normalized and added to the outcome advantage, while the verifier reward, policy-ratio construction, clipping rule, and KL regularizer remain unchanged.
  }
  \label{fig:overview}

\end{figure}

\section{Methodology}
\label{sec:method}

\subsection{Preliminary: Sparse Autoencoders}
\label{sec:prelim-sae}

Sparse Autoencoders (SAEs) decompose dense model activations into sparse latent
features. Given an activation vector $\mathbf{x}\in\mathbb{R}^{d}$ from a
target layer, an SAE encoder maps it to an overcomplete sparse code
$\mathbf{f}={f_{\mathrm{enc}}}(\mathbf{x})\in\mathbb{R}^{n}$, typically with $n\gg d$, and a
decoder reconstructs the original activation. Standard SAEs can be written as
\begin{equation}
\mathbf{f}=f_{\mathrm{enc}}(\mathbf{x})
=\sigma\!\left(\mathbf{W}_{\mathrm{enc}}(\mathbf{x}-\mathbf{b})\right),
\qquad
\hat{\mathbf{x}}=f_{\mathrm{dec}}(\mathbf{f})
=\mathbf{W}_{\mathrm{dec}}\mathbf{f}+\mathbf{b},
\end{equation}
where $\mathbf{W}_{\mathrm{enc}}\in\mathbb{R}^{n\times d}$,
$\mathbf{W}_{\mathrm{dec}}\in\mathbb{R}^{d\times n}$, $\mathbf{b}$ is a
centering bias, and $\sigma$ is an activation or sparsification function. The
SAE is trained to reconstruct activations while encouraging sparse codes:
\begin{equation}
    \mathcal{L}_{\mathrm{SAE}}(\mathbf{x})
    =
    \|\mathbf{x}-\hat{\mathbf{x}}\|_2^2
    +
    \lambda\,\mathcal{S}({f_{\mathrm{enc}}}(\mathbf{x})).
\end{equation}
Here $\mathcal{S}$ is commonly chosen as an $\ell_1$ penalty on the sparse code
$\mathbf{f}$. In this work, we use a BatchTopK SAE \citep{bussmann2024batchtopk},
which imposes hard sparsity by selecting the largest latent activations across
a minibatch. This allows different samples to activate different numbers of features while controlling the average number of active latents. The SAE is trained before policy optimization and kept frozen throughout RL training. We use it only as a sparse representation space for constructing token-level credit signals.

\subsection{Preliminary: Outcome-Supervised Policy Optimization}
\label{sec:prelim-rl}

We consider outcome-supervised RL training for reasoning tasks. For each
question \(q\), the policy samples \(G\) rollouts
\(\{o_{qg}\}_{g=1}^{G}\), where \(g\) indexes the rollout sampled for the same
question, and receives a scalar verifier reward \(r_{qg}\) for each completed
rollout. Let \(\mathcal{J}_{qg}\) denote the set of positions of the generated response tokens; this is used in the policy objective for rollout \(o_{qg}\).
Accordingly, \(|\mathcal{J}_{qg}|\) is the generated response length used for the objective.

\paragraph{Outcome Advantage.}
The rollout-level outcome advantage is computed by normalizing rewards within
each question group, following the group-relative normalization used in
GRPO-style RLVR training \citep{shao2024deepseekmathpushinglimitsmathematical}:
\begin{equation}
A^{\mathrm{out}}_{qg}
=
\begin{cases}
0, & \sigma_q^r < \epsilon_{\mathrm{std}},\\[1mm]
\tfrac{r_{qg}-\mu_q^r}{\sigma_q^r+\epsilon_{\mathrm{std}}}, & \text{otherwise},
\end{cases}
\label{eq:outcome-adv}
\end{equation}
where \(\mu_q^r\) and \(\sigma_q^r\) are the mean and standard deviation of
\(\{r_{q1},\ldots,r_{qG}\}\), and \(\epsilon_{\mathrm{std}}\) is a small
numerical constant. In typical outcome-supervised training (including GRPO and GSPO), this scalar
advantage is broadcast to every generated token:
$
A_{qg,t}=A^{\mathrm{out}}_{qg}, t\in\mathcal{J}_{qg}.
$
This provides group-relative credit across rollouts but does not distinguish
the contributions of individual tokens within a rollout.

\textbf{GRPO.}
GRPO can be viewed as a critic-free variant of PPO-style clipped policy
optimization, where the learned value baseline is replaced by group-relative
reward normalization \citep{schulman2017proximalpolicyoptimizationalgorithms,shao2024deepseekmathpushinglimitsmathematical}.
It uses token-level policy ratios in the clipped surrogate:
\begin{equation}
r_{qg,t}(\theta)
=
\tfrac{
\pi_\theta(o_{qg,t}\mid q,o_{qg,<t})
}{
\pi_{\theta_{\mathrm{old}}}(o_{qg,t}\mid q,o_{qg,<t})
}.
\label{eq:grpo-token-ratio}
\end{equation}
With the per-token advantage \(A_{qg,t}\), the GRPO objective can be written as
\begin{align}
\mathcal{L}_{\mathrm{GRPO}}(\theta)
=&
-\mathbb{E}_{q,g,t\in\mathcal{J}_{qg}}
\left[
\min\left(
r_{qg,t}(\theta) A_{qg,t},
\mathrm{clip}\!\left(
r_{qg,t}(\theta),
1-\epsilon_{\mathrm{clip}},
1+\epsilon_{\mathrm{clip}}
\right) A_{qg,t}
\right)
\right] \nonumber
\\
&+
\beta\,\mathrm{KL}\!\left[\pi_\theta\Vert\pi_{\mathrm{ref}}\right].\label{eq:grpo}
\end{align}
\textbf{GSPO.}
GSPO replaces token-level importance ratios with a length-normalized
sequence-level likelihood ratio and applies clipping at the sequence level \citep{zheng2025gspo}.
For rollout \(o_{qg}\), the sequence ratio is
\begin{equation}
s_{qg}(\theta)
=
\exp\left(
\tfrac{1}{|\mathcal{J}_{qg}|}
\sum\nolimits_{t\in\mathcal{J}_{qg}}
\log
\frac{
\pi_\theta(o_{qg,t}\mid q,o_{qg,<t})
}{
\pi_{\theta_{\mathrm{old}}}(o_{qg,t}\mid q,o_{qg,<t})
}
\right).
\label{eq:gspo-seq-ratio}
\end{equation}
With a rollout-level advantage \(A_{qg}\), the original GSPO surrogate can be written as
\begingroup\makeatletter\def\f@size{9.5}\check@mathfonts
\def\maketag@@@#1{\hbox{\m@th\normalsize\normalfont#1}}%
\begin{align}
\mathcal{L}_{\mathrm{GSPO}}(\theta)
=&
-\mathbb{E}_{q,g}
\left[
\min\left(
s_{qg}(\theta) A_{qg},
\mathrm{clip}\!\left(
s_{qg}(\theta),
1-\epsilon_{\mathrm{clip}},
1+\epsilon_{\mathrm{clip}}
\right) A_{qg}
\right)
\right] +
\beta\,\mathrm{KL}\!\left[\pi_\theta\Vert\pi_{\mathrm{ref}}\right].
\label{eq:gspo}
\end{align}
\endgroup
Since \modelname{} supplies token-specific advantages \(A_{qg,t}\), we use the
token-wise GSPO variant introduced for advantage customization \citep{zheng2025gspo}.
This variant preserves the same sequence-level ratio value and clipping condition,
but writes the loss over token positions so that \(A_{qg,t}\) can be substituted
directly. We provide the full token-wise GSPO surrogate in
Appendix~\ref{app:gspo_token}. %

In vanilla GRPO and GSPO, \(A_{qg,t}=A^{\mathrm{out}}_{qg}\) for all
\(t\in\mathcal{J}_{qg}\). In \modelname{}, we replace it with a shaped
token-level advantage, while leaving the verifier reward, policy-ratio  construction, clipping rule, and KL regularizer unchanged.

\subsection{Plug-in Outcome-Supervised Policy Optimization with \modelname{}}
\label{sec:saepo-method}

Our \modelname{} is a plug-in advantage-shaping module for outcome-supervised policy
optimization. \figref{fig:overview} shows an overview of \modelname{}. It constructs a
question-local token-level semantic signal from SAE features of the
current policy's rollouts, then adds the normalized signal to the standard
outcome advantage. The verifier reward, policy-ratio construction, clipping
rule, and KL regularizer are left unchanged, while advantage shaping changes the policy update. Appendix \ref{app:theory} provides an on-policy analysis of the expected SAE update through a local surrogate and relates it to a question-weighted success-rate gradient plus an error term.

\textbf{Step 1: Split Rollouts by Outcome Correctness.}
For each question \(q\), the current policy samples \(G\) rollouts
\(\{o_{qg}\}_{g=1}^{G}\), where \(g\) indexes the rollout. After grading the
final answer of each rollout, we split the group into correct and incorrect
rollout sets:
\begin{equation}
y_{qg}
=
\mathbf{1}\{\text{the final answer of }o_{qg}\text{ is correct}\},
\qquad
\mathcal{C}_q=\{g:y_{qg}=1\},\quad
\mathcal{I}_q=\{g:y_{qg}=0\}.
\label{eq:correct-split}
\end{equation}
We use this split to construct the SAE feature contrast. If either
\(\mathcal{C}_q\) or \(\mathcal{I}_q\) is empty, the question has no
correct-versus-incorrect contrast and receives no SAE signal for that update.

\textbf{Step 2: Encode Token Hidden States into Sparse Features.}
For each generated-token position \(t\in\mathcal{J}_{qg}\), let
\(H_{qg,t}\in\mathbb{R}^{d}\) denote the target-layer hidden state corresponding to the policy-loss token. We encode this hidden state with the pretrained SAE:
\begin{equation}
\phi_{qg,t}={f_{\mathrm{enc}}}(H_{qg,t})\in\mathbb{R}^{n},
\label{eq:sae-feature}
\end{equation}
{which generates SAE features for later semantic alignment.}

\textbf{Step 3.1: {Semantic Alignment across Rollouts}.}
The core of our method is {comparing each token within a rollout against other rollouts of its own question group}, rather than unrelated questions.
{Since different rollouts of the same question may reach the same intermediate result at different positions, we do not require tokens to align by position.} {For token $t$ in rollout $g$ of question $q$, we compare its SAE feature with those of every generated token $u$ in another rollout $h\ne g$ of the same question. A softmax with temperature $\tau>0$ assigns larger weights to tokens with higher cosine similarity. The weighted sum of these similarities gives one score for rollout $h$:}
\begin{equation}
\omega^{(h)}_{qg,t,u}
=
\tfrac{\exp\!\big(\cos(\phi_{qg,t},\phi_{qh,u})/\tau\big)}
     {\sum_{u'\in\mathcal{J}_{qh}}\exp\!\big(\cos(\phi_{qg,t},\phi_{qh,u'})/\tau\big)},
\quad
m^{(h)}_{qg,t}
=
\sum\nolimits_{u\in\mathcal{J}_{qh}}
\omega^{(h)}_{qg,t,u}\,
\cos(\phi_{qg,t},\phi_{qh,u}).
\label{eq:soft-align}
\end{equation}
Repeating this comparison for each rollout $h\ne g$ yields one score $m^{(h)}_{qg,t}$ for every other rollout $h$.
We then average these scores separately over the correct and incorrect rollouts:
\begin{equation}
s^{+}_{qg,t}
=
\tfrac{1}{|\mathcal{C}_q\setminus\{g\}|}
\sum\nolimits_{h\in\mathcal{C}_q\setminus\{g\}} m^{(h)}_{qg,t},
\qquad
s^{-}_{qg,t}
=
\tfrac{1}{|\mathcal{I}_q\setminus\{g\}|}
\sum\nolimits_{h\in\mathcal{I}_q\setminus\{g\}} m^{(h)}_{qg,t}.
\label{eq:side-scores}
\end{equation}
{We exclude $g$ to avoid matching token $t$ to itself. The resulting $s^+_{qg,t}$ and $s^-_{qg,t}$ measure its average similarity to correct and incorrect rollouts of the same question, respectively. If either set in \eqnref{eq:side-scores} is empty, rollout $g$ receives no SAE score.}

\textbf{Step 3.2: Calculate Token-Level Semantic Score.}
{We then compute the difference of two alignment scores from Step 3.1 and scale it by the balance of correct and incorrect rollouts for question $q$:}
\begin{equation}
S_{qg,t}
=
\rho_q
\left[
s^{+}_{qg,t}
-
s^{-}_{qg,t}
\right],
\qquad
t\in\mathcal{J}_{qg},
\label{eq:raw-score}
\end{equation}
{We call $S_{qg,t}$ the \emph{semantic score}. Here $\rho_q=2\min(|\mathcal{C}_q|,|\mathcal{I}_q|)/(|\mathcal{C}_q|+|\mathcal{I}_q|)$ downweights imbalanced correct/incorrect splits. A positive $S_{qg,t}$ means more token $t$'s similar tokens are in correct rollouts than in incorrect rollouts, and therefore token $t$ aligns more strongly with correct than incorrect rollouts of the same question; a negative value means the reverse.}

\textbf{Step 4: Advantage Shaping with Token-Level Semantic Score.}
The raw semantic scores have
batch-dependent scale, so we standardize them before adding them to the outcome
advantage. Specifically, for each update batch, we compute the mean \(\mu_S\)
and standard deviation \(\sigma_S\) {over all defined raw scores \(S_{qg,t}\)}.
{For these tokens}, we set
\[
\widehat{S}_{qg,t}
=
\tfrac{S_{qg,t}-\mu_S}{\sigma_S+\epsilon_{\mathrm{score}}},
\]
where \(\epsilon_{\mathrm{score}}\) is a small numerical constant for stability.
For unscored tokens, we set \(\widehat{S}_{qg,t}=0\).
{If no scores are defined in a batch, all tokens receive $\widehat{S}_{qg,t}=0$.} The final shaped advantage is
\begin{equation}
A_{qg,t}
=
A^{\mathrm{out}}_{qg}
+
\alpha\,\widehat{S}_{qg,t},
\qquad t\in\mathcal{J}_{qg},
\label{eq:final-adv}
\end{equation}
where \(\alpha\) controls the strength of the SAE signal. When \(\alpha=0\),
the update reduces to the vanilla outcome-only update. {Tokens in unscored
rollouts also use the outcome-only advantage.}

\textbf{Plug-in Policy Optimization.}
We instantiate \modelname{} as GRPO + SAEPO and GSPO + SAEPO by substituting
\eqnref{eq:final-adv} into the corresponding objective (\eqnref{eq:grpo} and \eqnref{eq:app-gspo-token-surrogate}). In both cases, the
only change from the vanilla optimizer is replacing the broadcast \emph{rollout-level} (sequence-level)
advantage \(A^{\mathrm{out}}_{qg}\) with the shaped \emph{token-level} advantage
\(A_{qg,t}\). The verifier reward, policy-ratio construction, clipping rule,
and KL regularizer are unchanged. The SAE is used only during fine-tuning to
compute \(\widehat{S}_{qg,t}\), and the final policy requires no SAE computation
at inference time.

\section{Experiments}
\label{sec:experiments}

In this section, we evaluate whether SAEPO's token-level credit improves outcome-supervised RLVR. %

\subsection{Experimental Setup}
\label{sec:experimental-setup}

\textbf{Training Data and Reward.}
For {mathematical reasoning}, all models are trained on the
\texttt{DeepScaleR-\allowbreak{}Preview-\allowbreak{}Dataset} {\citep{deepscaler2025}}, which contains around
\(40\)K math problem-answer pairs from competition-style sources such as AIME,
AMC, Omni-MATH, {and} STILL.
For each question, the policy samples a group of rollouts and receives an
outcome reward from the final answer.
The reward combines a binary correctness score from \texttt{math\_verify}
with a small format bonus for producing a boxed answer.
For code generation, we train on the TACO, PrimeIntellect SYNTHETIC-1,
and LiveCodeBench-v5 training splits of the \texttt{DeepCoder-\allowbreak{}Preview-\allowbreak{}Dataset} \citep{deepcoder2025}.

\textbf{Backbones and Baselines.}
We use \texttt{Qwen3-\allowbreak{}1.7B-\allowbreak{}Base} and \texttt{Qwen3-\allowbreak{}4B-\allowbreak{}Base} {\citep{yang2025qwen3technicalreport}}
{and \texttt{OLMo-\allowbreak{}3-\allowbreak{}7B-\allowbreak{}Base} \citep{olmo2025olmo}} as policy backbones.
For each backbone, we compare GRPO + \modelname{} against vanilla \textbf{GRPO}, and
GSPO + \modelname{} against vanilla \textbf{GSPO}.
We also include three state-of-the-art token-level credit-assignment methods as baselines: \textbf{KTAE} \citep{sun2025ktae}, \textbf{THR} \citep{deng2025tokenhidden}, and \textbf{GTPO} \citep{tan2025gtpo}.

\textbf{Evaluation.}
We evaluate accuracy on six mathematical reasoning benchmarks: MATH-500 {\citep{lightman2023letsverifystepstep}},
GSM8K {\citep{cobbe2021trainingverifierssolvemath}}, AIME25 {\citep{aime25}},
AMC {\citep{numina_math_datasets}}, Minerva Math {\citep{lewkowycz2022solvingquantitativereasoningproblems}},
and OlympiadBench {\citep{he2024olympiadbenchchallengingbenchmarkpromoting}}.
{For code generation, we report LiveCodeBench-v5 \citep{jain2024livecodebench,deepcoder2025} and
LeetCode Dataset v0.3.1 \citep{xia2025leetcodedatasettemporaldatasetrobust}
(228 held-out test problems), executing every unit test and
counting a problem as solved only if all tests pass.} {More implementation details are in ~\appref{app:training_details} and~\appref{app:evaluation_details}.}

\begin{table}[!t]
\centering
\caption{
Accuracy (\%) on six math and two code benchmarks across three LLM backbones, with unweighted averages. \textbf{Bold} marks the best result within each backbone's GRPO or GSPO block.
}
\vspace{0.05cm}

{
\setlength{\tabcolsep}{3.2pt}
\resizebox{\linewidth}{!}{%
\begin{tabular}{@{}l@{\hspace{10pt}}lcccccccccc@{}}

\toprule
& & \multicolumn{7}{c}{\textbf{Math}} & \multicolumn{3}{c}{\textbf{Code}} \\
\cmidrule(lr){3-9}\cmidrule(lr){10-12}

\textbf{Backbone}
& \textbf{Method}
& \textbf{MATH-500}
& \textbf{GSM8K}
& \textbf{AIME25}
& \textbf{AMC}
& \textbf{Minerva}
& \textbf{Olympiad}
& \textbf{Avg.}
& \textbf{LCB-v5}
& \textbf{LeetCode}
& \textbf{Avg.} \\

\midrule

& Vanilla GRPO
& 63.2 & 74.9 & \textbf{3.3} & 26.5 & 23.9 & 21.5
& 35.56 & 4.66 & \textbf{12.28} & 8.47 \\

& GRPO + KTAE
& 61.6 & 76.1 & 0.0 & 22.9 & 27.6 & 20.2
& 34.73 & 5.38 & 11.84 & 8.61 \\

& GRPO + THR
& 62.8 & \textbf{76.4} & \textbf{3.3} & 27.7 & 26.8 & \textbf{23.3}
& 36.72 & 5.73 & 7.02 & 6.38 \\

& GRPO + GTPO
& 45.2 & 74.0 & \textbf{3.3} & 13.3 & 19.9 & 9.3
& 27.50 & 3.58 & 4.82 & 4.20 \\

\rowcolor{gray!10}
\cellcolor{white}
& GRPO + SAEPO (Ours)
& \textbf{65.6} & 76.3 & \textbf{3.3} & \textbf{34.9}
& \textbf{29.8} & 22.1 & \textbf{38.68}
& \textbf{6.81} & 11.40 & \textbf{9.11} \\

\cmidrule(l){2-12}

& Vanilla GSPO
& 62.8 & 76.9 & 6.7 & 27.7 & 23.9 & 21.8
& 36.63 & \textbf{5.73} & 9.65 & 7.69 \\

& GSPO + KTAE
& 60.4 & 75.0 & 0.0 & 22.9 & 27.2 & 19.3
& 34.13 & 4.66 & 10.09 & 7.37 \\

& GSPO + THR
& 64.6 & 77.0 & \textbf{10.0} & 24.1 & 26.5 & 20.6
& 37.12 & 4.66 & 8.33 & 6.50 \\

& GSPO + GTPO
& 36.8 & 73.2 & 0.0 & 9.6 & 17.6 & 7.3
& 24.09 & 1.43 & 7.02 & 4.23 \\

\rowcolor{gray!10}
\cellcolor{white}
\multirow{-10}{*}{\textit{Qwen3-1.7B-Base}}
& GSPO + SAEPO (Ours)
& \textbf{65.8} & \textbf{78.7} & 6.7 & \textbf{32.5}
& \textbf{30.5} & \textbf{23.1} & \textbf{39.55}
& 5.02 & \textbf{11.40} & \textbf{8.21} \\

\midrule

& Vanilla GRPO
& 77.6 & 89.5 & \textbf{10.0} & 44.6 & 36.0 & 32.9
& 48.43 & \textbf{13.26} & 10.96 & 12.11 \\

& GRPO + KTAE
& 75.6 & 88.5 & 3.3 & 45.8 & \textbf{41.2} & 32.0
& 47.74 & \textbf{13.26} & 12.72 & 12.99 \\

& GRPO + THR
& 77.0 & 88.9 & 6.7 & 42.2 & 32.0 & 33.7
& 46.74 & 11.47 & 11.40 & 11.44 \\

& GRPO + GTPO
& 53.2 & 87.9 & 0.0 & 26.5 & 24.6 & 12.9
& 34.19 & 6.81 & 5.26 & 6.04 \\

\rowcolor{gray!10}
\cellcolor{white}
& GRPO + SAEPO (Ours)
& \textbf{78.0} & \textbf{90.8} & \textbf{10.0} & \textbf{47.0}
& 39.7 & \textbf{34.3} & \textbf{49.97}
& 12.54 & \textbf{22.81} & \textbf{17.68} \\

\cmidrule(l){2-12}

& Vanilla GSPO
& 77.0 & 90.0 & 6.7 & \textbf{43.4} & 36.4 & 32.9
& 47.73 & 12.90 & 13.16 & 13.03 \\

& GSPO + KTAE
& 73.2 & 89.3 & \textbf{10.0} & 39.8 & 39.0 & 30.6
& 46.97 & 11.47 & 8.33 & 9.90 \\

& GSPO + THR
& 74.4 & 88.8 & \textbf{10.0} & 42.2 & 34.2 & 33.8
& 47.23 & 11.11 & 12.28 & 11.70 \\

& GSPO + GTPO
& 47.6 & 85.7 & 0.0 & 18.1 & 23.5 & 12.2
& 31.17 & 6.81 & 5.26 & 6.04 \\

\rowcolor{gray!10}
\cellcolor{white}
\multirow{-10}{*}{\textit{Qwen3-4B-Base}}
& GSPO + SAEPO (Ours)
& \textbf{78.4} & \textbf{90.1} & \textbf{10.0} & \textbf{43.4}
& \textbf{43.8} & \textbf{35.2} & \textbf{50.15}
& \textbf{14.34} & \textbf{18.86} & \textbf{16.60} \\

\midrule

& Vanilla GRPO
& 55.2 & 86.6 & 0.0 & 25.3 & 24.6 & 20.0
& 35.29 & 8.60 & 7.89 & 8.25 \\

& GRPO + KTAE
& 62.0 & 86.6 & 3.3 & 28.9 & \textbf{29.4} & 22.3
& 38.75 & 8.96 & 7.89 & 8.43 \\

& GRPO + THR
& 60.0 & 87.3 & 0.0 & 28.9 & 23.2 & 22.4
& 36.96 & \textbf{9.68} & 7.89 & 8.79 \\

& GRPO + GTPO
& 41.0 & 85.4 & 0.0 & 18.1 & 19.9 & 9.5
& 28.98 & 4.30 & 3.95 & 4.12 \\

\rowcolor{gray!10}
\cellcolor{white}
& GRPO + SAEPO (Ours)
& \textbf{65.4} & \textbf{88.3} & \textbf{6.7} & \textbf{37.3}
& 26.8 & \textbf{27.0} & \textbf{41.92}
& \textbf{9.68} & \textbf{10.96} & \textbf{10.32} \\

\cmidrule(l){2-12}

& Vanilla GSPO
& 67.2 & \textbf{86.0} & \textbf{10.0} & \textbf{36.1}
& 25.0 & 26.0 & 41.71
& \textbf{11.11} & 9.21 & 10.16 \\

& GSPO + KTAE
& 63.6 & 85.5 & 6.7 & 32.5 & 25.7 & 26.7
& 40.13 & 8.60 & 8.33 & 8.47 \\

& GSPO + THR
& 60.8 & 85.5 & \textbf{10.0} & 27.7 & 25.0 & 20.5
& 38.25 & 9.68 & 8.33 & 9.01 \\

& GSPO + GTPO
& 50.4 & 84.8 & 0.0 & 21.7 & 18.8 & 9.6
& 30.88 & 4.66 & 5.70 & 5.18 \\

\rowcolor{gray!10}
\cellcolor{white}
\multirow{-10}{*}{\textit{OLMo-3-7B-Base}}
& GSPO + SAEPO (Ours)
& \textbf{68.6} & 85.6 & 6.7 & \textbf{36.1}
& \textbf{30.1} & \textbf{31.3} & \textbf{43.07}
& 9.68 & \textbf{14.47} & \textbf{12.08} \\

\bottomrule
\end{tabular}%
}
}

\label{tab:benchmark}

\end{table}

\begin{figure}[!t]
  \centering
  \includegraphics[width=\linewidth]
  {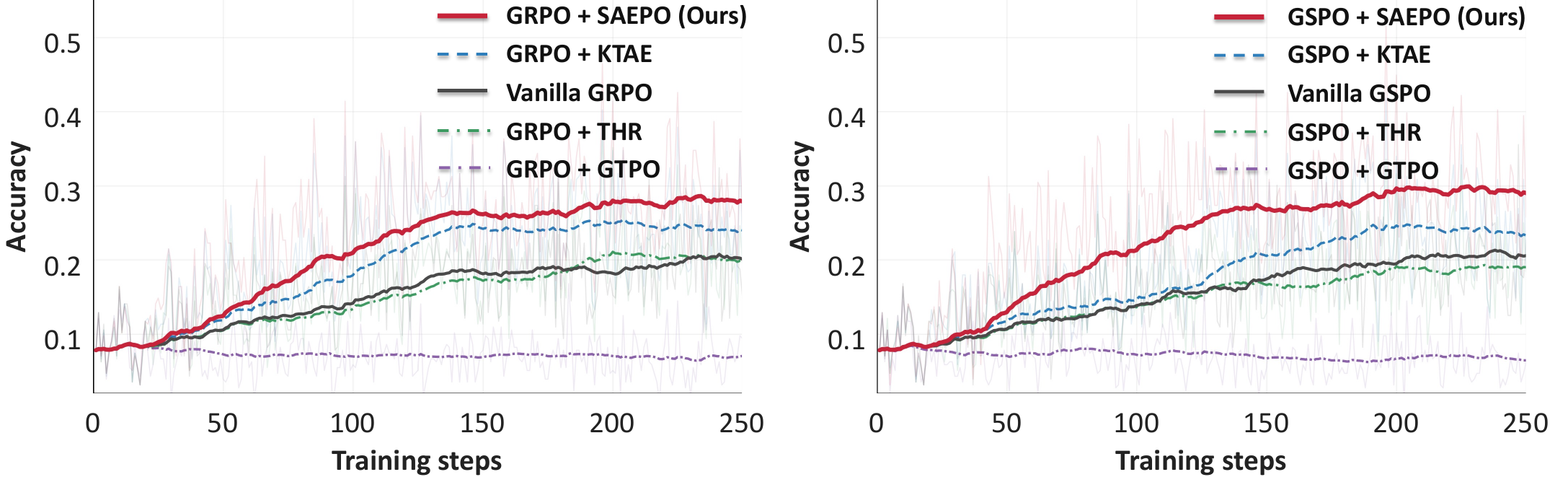}

 \caption{Training dynamics on {\texttt{Qwen3-\allowbreak{}1.7B-\allowbreak{}Base}}.
  We plot training accuracy curves during RL \emph{\textbf{training}} for GRPO and GSPO,
  comparing each vanilla policy optimization algorithm with its \modelname{}-augmented variant
  {and with credit-assignment baselines (i.e., KTAE, THR, and GTPO)}.
  \modelname{} shows faster convergence in training accuracy and generally reaches higher training accuracy.
  }
   \label{fig:dynamics}

\end{figure}

\begin{figure}[!t]
  \centering
  \includegraphics[width=\linewidth]
  {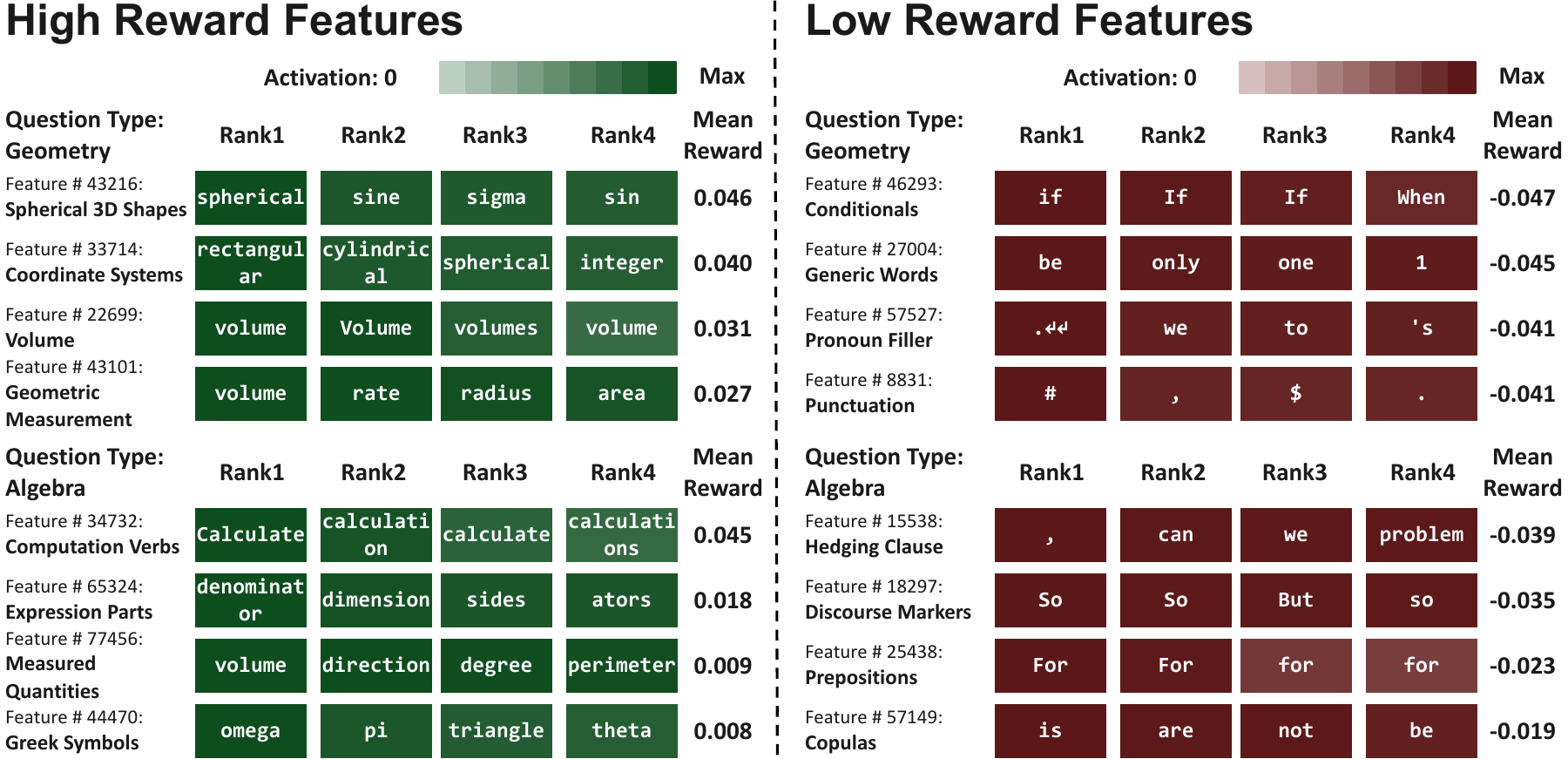}

   \caption{
      SAE features (concepts) associated with the highest and the lowest \modelname{}
      token scores, computed from rollouts on the ``Algebra'' and ``Geometry'' subjects
      in MATH-500. Feature-level rewards are
      obtained by averaging semantic scores over tokens where each SAE feature is active.
      Rows show features with highest/lowest rewards and their top activating tokens;
      color intensity indicates normalized activation within each feature{.}
      \textbf{Left:} High-reward features correspond to subject-relevant concepts{, such as spherical and coordinate vocabulary in Geometry, and computation and
      expression terms in Algebra}. \textbf{Right:} Low-reward features mostly capture
      {connectives, copulas,} punctuation, stop words, and discourse markers. These
      results suggest that the \modelname{}-derived signal concentrates on semantically
      meaningful reasoning features.
}
   \label{fig:feature_analysis}

\end{figure}

\subsection{Results}
\label{sec:results}

\textbf{Main Results.}
{\tabref{tab:benchmark} compares \modelname{} with vanilla GRPO and GSPO, as well as three credit-assignment baselines, i.e., KTAE, THR, and GTPO, across all three LLM backbones and both task types.}
On \texttt{Qwen3-\allowbreak{}1.7B-\allowbreak{}Base}, \modelname{} successfully enhances both policy optimization algorithms, improving GRPO from {\(35.56\%\) to \(38.68\%\)} and GSPO from {\(36.63\%\) to \(39.55\%\)}.
On \texttt{Qwen3-\allowbreak{}4B-\allowbreak{}Base}, \modelname{} improves the {six-benchmark math average} from {\(48.43\%\) to \(49.97\%\)} with GRPO, and from {\(47.73\%\) to \(50.15\%\)} with GSPO.
{Importantly, \modelname{} also consistently outperforms all three credit-assignment baselines in average accuracy across every backbone, policy algorithm, and task type.}
{For example, on \texttt{OLMo-\allowbreak{}3-\allowbreak{}7B-\allowbreak{}Base}, \modelname{} achieves \(41.92\%\) and \(43.07\%\) average math accuracy with GRPO and GSPO, respectively, compared with \(38.75\%\) and \(40.13\%\) for the strongest credit-assignment baseline.}
The same advantage extends to code generation benchmarks.
These results show that \modelname{}'s token rewards consistently (1) improve the vanilla GRPO/GSPO across {all three backbones and both task types}, and (2) outperform existing credit-assignment approaches.

\textbf{Training Efficiency.}
\figref{fig:dynamics} compares training accuracy curves on {\texttt{Qwen3-\allowbreak{}1.7B-\allowbreak{}Base}}, {including vanilla GRPO/GSPO and the credit-assignment baselines KTAE, THR, and GTPO}.
For both GRPO and GSPO, \modelname{} improves early-stage accuracy growth and generally reaches higher accuracy under the same training budget.
{It also enjoys faster convergence and generally higher training accuracy than the three credit-assignment baselines throughout training.}
This suggests that the \modelname{}-derived \emph{token-level} signal provides a more informative advantage estimate than broadcasting the same \emph{rollout-level} outcome advantage to all generated tokens, while also providing more effective credit assignment than existing token-level alternatives.

\textbf{Interpretability of the Credit Signal.}
\figref{fig:feature_analysis} visualizes \modelname{}'s SAE features (concepts) that have the highest and lowest average token-level {semantic scores} on Algebra and Geometry rollouts from MATH-500. For each top feature, we list its top-$4$ tokens with the highest activation in the SAE.
High-reward features correspond to subject-relevant mathematical concepts, such as {computation and expression terms}, circle/sphere concepts, and geometric measurements.
In contrast, low-reward features mainly capture {connectives}, punctuation, generic stop words, {copulas}, pronouns, and discourse markers.
This suggests that the positive credit assigned by \modelname{} tends to concentrate on sparse features tied to question-relevant content.

\begin{figure}[t]

  \centering
  \includegraphics[width=\linewidth]{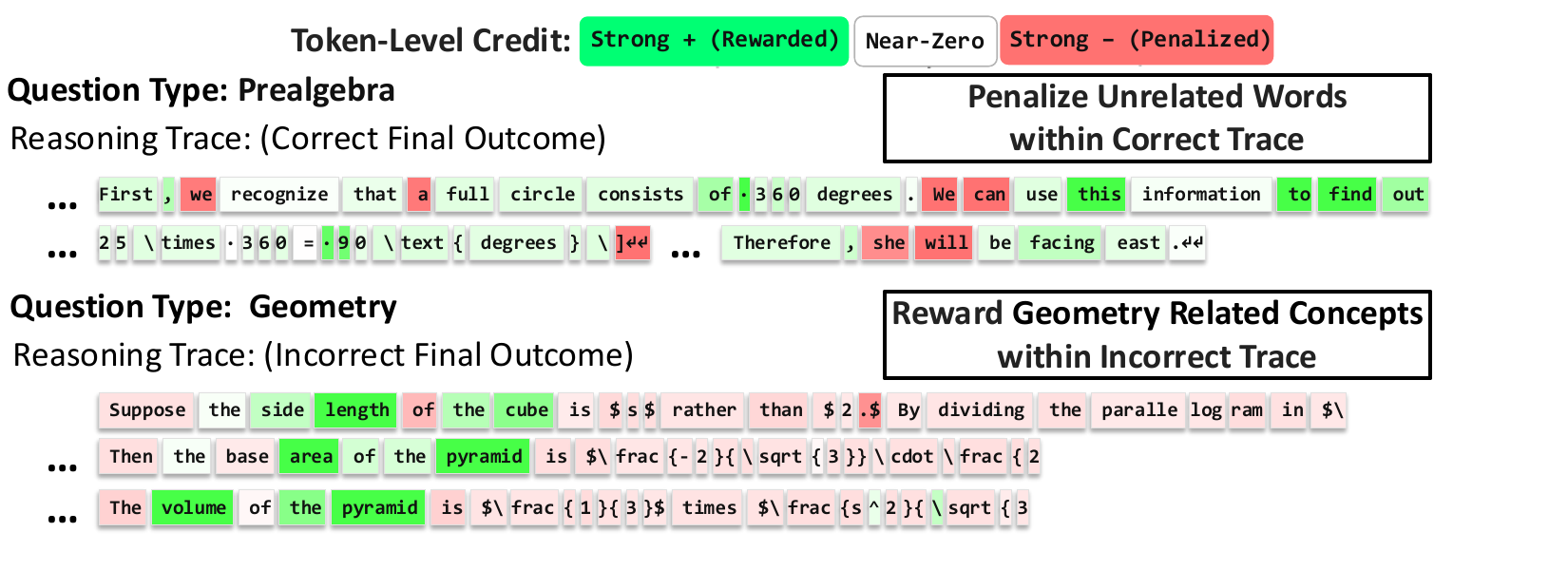}

  \caption{Examples {of how} \modelname{} can assign {a} fine-grained credit signal {to} each token. For rollouts from different types of questions, we visualize the {\modelname{}-derived} final shaped advantages $A_{qg,t}=A^{\mathrm{out}}_{qg}+\alpha\widehat{S}_{qg,t}$ from \eqnref{eq:final-adv} assigned to individual tokens in the rollouts. \green{Green tokens} represent positive {credit} that {rewards} the tokens, \gray{gray tokens} represent near-zero credit, while \red{red tokens} represent negative credit {that penalizes} the tokens. \modelname{} tends to reward {\emph{semantically meaningful} tokens like \emph{question-relevant tokens} in incorrect trajectories and \emph{reasoning} tokens in correct trajectories}, while penalizing \emph{uninformative} or \emph{misleading tokens} {even when they are in correct trajectories}, achieving fine-grained token-level credit assignment over the reasoning trace.}
  \label{fig:Case Study}

\end{figure}

{\figref{fig:Case Study} provides a token-level view of \modelname{}.
Importantly, SAEPO's signal is not simply a copy of the final rollout outcome.
In a correct trace for {a Prealgebra} problem (\figref{fig:Case Study}(top)), \modelname{} still assigns negative final advantages to unrelated stop words (e.g., {``we''}), while rewarding tokens involved in the reasoning (e.g., {``circle'' and ``degrees''}).
Conversely, in an incorrect trace for {a} geometry question (\figref{fig:Case Study}(bottom)), the {rollout} receives negative outcome supervision, but some reasoning tokens
still contain relevant concepts (e.g., {``length'', ``area'', ``volume'', and ``pyramid''}) and receive positive final advantages.
This illustrates the main advantage of our \modelname{}'s token-level signal: it can distinguish useful and unhelpful reasoning components within the same trajectory, while a rollout-level advantage would assign the same credit to all tokens (more examples in~\appref{appsec:case_study}). }

\begin{table}[!t]
    \centering
    \caption{
    Ablation studies with \texttt{Qwen3-\allowbreak{}1.7B-\allowbreak{}Base} on GRPO with \modelname{}.
    We report accuracy~(\%) on the same six-benchmark evaluation suite as \tabref{tab:benchmark}. \textbf{Bold} marks the best result.
    }
    \label{tab:ablation_basis_sign}
    \vskip 0.1cm
    \resizebox{\linewidth}{!}{
    \begin{tabular}{lccccccc}
    \toprule
    \textbf{Method}  & \textbf{MATH-500} & \textbf{GSM8K}
        & \textbf{AIME25} & \textbf{AMC} & \textbf{Minerva}
        & \textbf{Olympiad} & \textbf{Avg.} \\
    \midrule
    Vanilla GRPO                     & 63.2 & 74.9 & \textbf{3.3} & 26.5 & 23.9 & 21.5 & 35.56 \\
    GRPO + Raw-Embedding Contrast &
        64.2   & 74.5 & 0.0  & 26.5 &30.1  & \textbf{23.9}  & 36.53  \\
    GRPO + Reverse \modelname{} Signal
        & 37.8  & 75.4 & 0.0   & 9.6 & 22.4  &7.6   & 25.48 \\
    \rowcolor{gray!10}
    GRPO + SAEPO (Ours)          & \textbf{65.6} & \textbf{76.3} & \textbf{3.3} & \textbf{34.9} & \textbf{29.8} & 22.1 & \textbf{38.68} \\
    \bottomrule
    \end{tabular}
    }

\end{table}

\textbf{Ablation Studies.}
\tabref{tab:ablation_basis_sign} shows the ablation study results for two core components of \modelname{} on \texttt{Qwen3-\allowbreak{}1.7B-\allowbreak{}Base} with GRPO.
Replacing SAE features with raw hidden states hurts performance, indicating that the sparse SAE basis is important for constructing a useful token-level signal.
Flipping the sign of the SAE-derived score leads to a large degradation, showing that the success-versus-failure direction carries meaningful training information.
The full GRPO + \modelname{} variant achieves the best average performance, {supporting the design of our method. More ablations are in~\appref{app:ablation_details}.}

\paragraph{Controlled Evaluation of Token-Level Scores.}

We conduct a paired minimal-edit experiment to examine the links between SAEPO's token-level semantic scores, local-step correctness, and prefix-continuation success rates. We sample 976 training problems and select 1,864 rollouts from \texttt{Qwen3-\allowbreak{}1.7B-\allowbreak{}Base}, comprising 904 correct and 960 incorrect rollouts. After removing rollouts with ambiguous step boundaries or steps that are difficult to edit, we retain 403 correct and 507 incorrect rollouts as candidates. For each candidate, \texttt{Qwen3-\allowbreak{}32B} selects one step and proposes a minimal edit: either introducing a single verifiable error into a correct step or correcting an error in an incorrect step. A separate blind judging pass of the same model evaluates both versions and confirms the intended change in step correctness, yielding 304 error introductions and 120 fixes for evaluation. All tokens outside the edited span remain unchanged.

To examine the link between semantic scores and local-step correctness, we feed the original and edited rollouts through the policy and recompute their semantic scores $S$, keeping the reference rollouts and outcome labels fixed. We define $\Delta S$ as the mean score over the edited tokens minus the mean over the original tokens they replace, divided by the standard deviation of scores across the original rollout. This normalization makes score changes comparable across rollouts.

\begin{table}[t]
    \centering
    \caption{Controlled changes in step correctness produce corresponding changes in semantic scores on average: introducing an error lowers the score, while fixing an error raises it.}
    \label{tab:controlled_step_edits}
    \normalsize
    \begin{tabular}{lcll}
        \toprule
        Edit
        & Samples
        & Example: original $\rightarrow$ edited
        & Mean $\Delta S$ [95\% CI] \\
        \midrule
        Introduce an error
        & 304
        & $8\times3=24 \;\rightarrow\; 8\times3=25$
        & $\mathbf{-0.58}$ [$-0.70$, $-0.47$] \\
        Fix an error
        & 120
        & Cylinder: $V=\frac{1}{3}\pi r^2h
          \;\rightarrow\; V=\pi r^2h$
        & $\mathbf{+0.16}$ [$+0.06$, $+0.25$] \\
        \bottomrule
    \end{tabular}

\end{table}

Table~\ref{tab:controlled_step_edits} shows that introducing an error decreases the score on average, whereas fixing an error increases it; both confidence intervals exclude zero. Thus, the semantic score responds to controlled changes in local reasoning correctness in the intended direction on average, supporting its use as a local credit signal.

To examine the link between local-step correctness and prefix-continuation success rates, we truncate each rollout immediately after the edited tokens and sample 32 continuations from each original prefix and 32 from its edited counterpart. Let $\widehat P$ denote the fraction of continuations with correct final answers, and let $\Delta\widehat P$ be the edited estimate minus the original estimate.

\begin{table}[t]
    \centering
    \caption{Controlled reasoning-step edits affect continuation success in the expected direction: introducing an error lowers the estimated success rate, while fixing an error raises it. These results support the relevance of the constructed edits to final-answer correctness.}
    \label{tab:controlled_continuations}
    \normalsize
    %
    \begin{tabular}{lcccc}
        \toprule
        Edit
        & Samples
        & Mean $\widehat P$: original $\rightarrow$ edited
        & Mean $\Delta\widehat P$ [95\% CI]
        & Lower / higher $\widehat P$ \\
        \midrule
        Introduce an error
        & 304
        & $0.74 \rightarrow 0.17$
        & $\mathbf{-0.57}$ [$-0.61$, $-0.54$]
        & \textbf{95\%} / 4\% \\
        Fix an error
        & 120
        & $0.15 \rightarrow 0.32$
        & $\mathbf{+0.17}$ [$+0.12$, $+0.22$]
        & 23\% / \textbf{62\%} \\
        \bottomrule
    \end{tabular}%
    %

\end{table}

As shown in Table~\ref{tab:controlled_continuations}, introducing an error lowers estimated success in 95\% of edits, reducing the mean from $0.74$ to $0.17$. Conversely, fixing an error raises estimated success in 62\% of edits, increasing the mean from $0.15$ to $0.32$. Thus, the controlled edits affect both local reasoning correctness and the policy's ability to reach a correct final answer. Together with Table~\ref{tab:controlled_step_edits}, these results show that semantic scores and continuation success respond in the same direction on average, providing empirical support for SAEPO's local credit signal.

\section{Conclusion}
We introduced \modelname{}, a sparse-autoencoder-guided framework for fine-grained token-level credit assignment in outcome-supervised RLVR. By contrasting correct and incorrect rollouts of the same question in a sparse SAE feature space, \modelname{} derives question-local semantic signals to reshape token-level advantages while preserving the verifier reward, policy-ratio construction, clipping rule, and KL regularizer. Across GRPO and GSPO, three LLM backbones, and math and code tasks, \modelname{} improves average accuracy over vanilla methods and credit-assignment baselines. Interpretability analyses show that the SAE-derived signal captures meaningful reasoning features. Future work includes extending the framework to other verifiable domains, SAE architectures, and optimization algorithms.



\bibliography{main}
\bibliographystyle{plainnat}

\newpage

\appendix
\raggedbottom
\section*{Appendix}

\section{Additional Method Details}
\label{app:method_details}

\subsection{Token-wise GSPO Surrogate}
\label{app:gspo_token}

For completeness, we describe the token-wise GSPO variant used in our
implementation. Recall that the original GSPO sequence-level ratio for
rollout \(o_{qg}\) is
\[
s_{qg}(\theta)
=
\exp\left(
\frac{1}{|\mathcal{J}_{qg}|}
\sum_{t\in\mathcal{J}_{qg}}
\log
\frac{
\pi_\theta(o_{qg,t}\mid q,o_{qg,<t})
}{
\pi_{\theta_{\mathrm{old}}}(o_{qg,t}\mid q,o_{qg,<t})
}
\right).
\]
To support token-specific advantages, we use the token-wise GSPO
construction introduced for advantage customization. Define the
effective token-wise weight as
\begin{equation}
\tilde{s}_{qg,t}(\theta)
=
\mathrm{sg}\!\left[s_{qg}(\theta)\right]\,
\frac{
\pi_\theta(o_{qg,t}\mid q,o_{qg,<t})
}{
\mathrm{sg}\!\left[\pi_\theta(o_{qg,t}\mid q,o_{qg,<t})\right]
},
\label{eq:app-gspo-token-weight}
\end{equation}
where \(\mathrm{sg}[\cdot]\) denotes the stop-gradient operator.
Numerically, \(\tilde{s}_{qg,t}(\theta)=s_{qg}(\theta)\), so the
clipping condition is still determined by the sequence-level ratio.
However, the gradient of each token loss term flows through the
corresponding token log-probability, allowing a token-specific
advantage \(A_{qg,t}\).

The token-wise GSPO surrogate is
\begin{align}
\mathcal{L}_{\mathrm{GSPO}}(\theta)
=&
-\mathbb{E}_{q,g,t\in\mathcal{J}_{qg}}
\left[
\min\!\left(
\tilde{s}_{qg,t}(\theta) A_{qg,t},
\mathrm{clip}\!\left(
\tilde{s}_{qg,t}(\theta),
1-\epsilon_{\mathrm{clip}},
1+\epsilon_{\mathrm{clip}}^{\mathrm{high}}
\right) A_{qg,t}
\right)
\right] \nonumber
\\
&+
\beta\,\mathrm{KL}\!\left[\pi_\theta\Vert\pi_{\mathrm{ref}}\right].
\label{eq:app-gspo-token-surrogate}
\end{align}
When all tokens in a rollout share the same advantage, this formulation
preserves the original GSPO sequence-level ratio value and clipping
behavior. In \modelname{}, we substitute the shaped token-level
advantage from Eq.~\eqref{eq:final-adv} into
Eq.~\eqref{eq:app-gspo-token-surrogate}.

\paragraph{Clip values used in this paper.}
Because the GSPO ratio is sequence-level and length-normalized, its
admissible clip range is much tighter than GRPO's. Following the GSPO
paper, all GSPO runs in this paper use asymmetric clipping
\(\epsilon_{\mathrm{clip}}=3{\times}10^{-4}\) and
\(\epsilon_{\mathrm{clip}}^{\mathrm{high}}=4{\times}10^{-4}\); GRPO uses
the symmetric \(\epsilon_{\mathrm{clip}}=0.2\) standard. The KL
coefficient \(\beta=0.02\) and reference policy \(\pi_{\mathrm{ref}}\)
are shared across all algorithms and arms.

\section{Theoretical Analysis}
\label{app:theory}

This appendix analyzes the token-level credit and advantage shaping in SAEPO. Let $P^\pi(s)$ be the probability that policy $\pi$, continuing from prefix $s$, reaches a correct answer. A token's local contribution is the change it produces in this probability. We first relate the raw score to similarity-weighted success (Lemma~1), then show how score changes enter the expected token-level update (Proposition~1). The connection to true local contribution depends on the representation, so we also test it through controlled edits and sampled continuations in Section~\ref{sec:results}. Preserving the loss form does not establish consistency with the original objective. We therefore further investigate advantage shaping: Proposition~2 identifies the local surrogate optimized by the SAE term, and Proposition~3 expresses this term as $\alpha$ times a question-weighted success-rate gradient plus a residual from score errors. The verifier reward, policy-ratio construction, clipping rule, and KL regularizer remain unchanged, while the advantage and hence the update change. The results below provide the full statements, assumptions, and proofs.

\subsection{Setup}\label{app:theory_setup}

\paragraph{Questions, Solutions and Success Probabilities.}
Questions $q$ are drawn from a fixed distribution with finite support (the training set). For a question $q$, the policy $\pi_\theta$ generates a solution $y=(y_1,\dots,y_T)$ token by token from a finite vocabulary $\mathcal{V}$. The length $T\le H$ is random: generation stops at the end-of-sequence token or at the maximum length $H$. The policy is a softmax policy whose probabilities are differentiable in $\theta$, so $\pi_\theta(a\mid s)>0$ for every token $a$ and prefix $s$. Let $s_t=(q,y_1,\dots,y_t)$ be the prefix after $t$ tokens, with $s_0=q$, and let $g_t=\nabla_\theta\log\pi_\theta(y_t\mid s_{t-1})$ for $t\le T$. For $t>T$ we set $s_t=s_T$ and set every per-token quantity ($g_t$, and $m_t$ and $\hat S_t$ below) to zero, so that $\sum_{t=1}^{T}=\sum_{t=1}^{H}$. The event $\{t\le T\}$ (the solution has not ended before step $t$) is determined by $s_{t-1}$, and $g_t$ is determined by $s_t$. Let $r(y)\in\{0,1\}$ indicate that $y$ is correct, and let $p_q=\Pr(r(y)=1)$ for $y\sim\pi_\theta(\cdot\mid q)$. For a prefix $s$ with $t$ tokens, let
\[
P^\pi(s)=\mathbb{E}\big[r(y)\mid s_t=s\big]
\]
be the probability that the policy reaches a correct answer when it continues from $s$. Thus $P^\pi(s_0)=p_q$, $P^\pi(s_T)=r(y)$, and $\mathbb{E}[P^\pi(s_t)\mid s_{t-1}]=P^\pi(s_{t-1})$ on $\{t\le T\}$.

\paragraph{Groups and Batches.}
A batch contains a fixed number $B$ of questions, each drawn from this distribution before any solution is generated, and, for each question, $G\ge 3$ solutions $y^{(1)},\dots,y^{(G)}$ sampled independently from $\pi_\theta(\cdot\mid q)$, so the batch has $M=BG$ solutions; $n_+$ and $n_-$ count the correct and incorrect solutions of a group. Every solution has $1\le T\le H$ tokens. All expectations are over batches sampled from the current policy.

\paragraph{Similarity.}
Let $z_t\in\mathbb{R}_{\ge 0}^{D}$ be the SAE code of the policy's hidden state at the $t$-th token of $y$, and let $z_u(y')$ be the code at the $u$-th token of another solution $y'$. The codes are nonnegative because the SAE applies a ReLU before its top-$k$ selection. The hidden state at token $t$ depends only on $s_t$; with a BatchTopK SAE, however, the sparsity threshold is computed over all tokens of a solution, so $z_t$ can also depend on the rest of $y$. With temperature $\tau>0$ and $c_u=\cos\big(z_t,z_u(y')\big)\in[0,1]$ (set to $0$ if either code is zero), the similarity between token $t$ of $y$ and another solution $y'$ is the softmax-weighted mean of these cosine similarities,
\[
k_t(y')=\sum_{u=1}^{|y'|}w_u\,c_u,\qquad
w_u=\frac{\exp(c_u/\tau)}{\sum_{u'=1}^{|y'|}\exp(c_{u'}/\tau)},
\]
so $0\le k_t(y')\le 1$. For a fixed solution $y'$, its average over the continuation of the prefix,
\[
k(s,y')=\mathbb{E}\big[k_t(y')\mid s_t=s\big],
\]
depends only on $s$ and $y'$, and $0\le k(s,y')\le 1$. If the code at a token depends only on its prefix (as for a TopK SAE), then $k_t(y')=k(s_t,y')$.

\paragraph{Scores.}
Consider the $i$-th solution $y=y^{(i)}$ of a group. Let $\mathcal{C}_i=\{j\ne i: r(y^{(j)})=1\}$ and $\mathcal{W}_i=\{j\ne i: r(y^{(j)})=0\}$ be the other correct and the other incorrect solutions. The tokens of $y$ are scored if both sets are nonempty (event $E_i$) and unscored otherwise. For a scored token, the raw score is
\[
m_t=\frac{1}{|\mathcal{C}_i|}\sum_{j\in\mathcal{C}_i}k_t\big(y^{(j)}\big)-\frac{1}{|\mathcal{W}_i|}\sum_{j\in\mathcal{W}_i}k_t\big(y^{(j)}\big),
\]
and $m_t=0$ for an unscored token. The SAE score is $S_t=\rho\,m_t$, with reliability weight $\rho=2\min(n_+,n_-)/G$. Scores are normalized over the scored tokens of the batch:
\[
\hat S_t=\frac{S_t-\mu_B}{\sigma_B+\epsilon_{\mathrm{score}}}\ \text{ for scored tokens},\qquad
\hat S_t=0\ \text{ for unscored tokens},
\]
where $\epsilon_{\mathrm{score}}>0$, and $\mu_B$ and $\sigma_B$ are the mean and sample standard deviation of $S_t$ over all scored tokens of the batch. A group with scored tokens has at least $G-1\ge 2$ scored solutions, so $\sigma_B$ is defined whenever the batch has a scored token; if it has none, all $\hat S_t=0$ (and we set $\mu_B=\sigma_B=0$).

\paragraph{Update.}
SAEPO replaces GRPO's outcome advantage $A$, which is shared by all tokens of a solution, by $A+\alpha\hat S_t$ with $\alpha>0$. We analyze one update on an on-policy batch. At $\theta=\theta_{\mathrm{old}}$ the importance ratios equal one and clipping is inactive, so the ascent direction (the negative gradient of the loss) is the sum of $\sum_t(A+\alpha\hat S_t)\,g_t$ over the solutions of the batch, divided by the loss normalizer. As in the implementation, $A$, $\hat S_t$ (including $\rho$, $\mu_B$, $\sigma_B$ and the SAE codes) and the normalizer are held fixed during differentiation. The KL term is the same as in GRPO and is not analyzed. Let $y$ be a solution chosen uniformly at random from the batch, and $q$ its question. We analyze the expected SAE term per solution,
\[
d(\theta)=\alpha\,\mathbb{E}\Big[\sum_{t=1}^{T}\hat S_t\,g_t\Big].
\]
The implemented loss is a token mean: it divides the sum over the batch by the number of tokens $N_B$ in the batch, which is random. Propositions~1--3 hold for any per-token score, so they apply exactly to the implemented update after replacing $\hat S_t$ by $(M/N_B)\,\hat S_t$, and $f$ below by $\tilde f(s)=\mathbb{E}\big[(M/N_B)\,\hat S_t\mid s_t=s\big]$; the random factor stays inside this conditional expectation. If $M/N_B$ is replaced by a deterministic constant $\beta>0$, as Corollary~1 does for $\rho$, $\mu_B$ and $\sigma_B$, the formula for $f$ in Corollary~1 and its coefficient $\lambda$ are multiplied by $\beta$. Since the question distribution has finite support, $\mathcal{V}$ is finite and $T\le H$, all expectations below are finite sums.

\paragraph{A Basic Identity.}
For every $t$ and every function $h$ of the prefix $s_{t-1}$,
\begin{equation}
\mathbb{E}\big[\mathbf{1}\{t\le T\}\,h(s_{t-1})\,g_t\big]=0.
\label{eq:app_theory_baseline}
\end{equation}
Indeed, on $\{t\le T\}$ the token $y_t$ is drawn from $\pi_\theta(\cdot\mid s_{t-1})$, and
$\mathbb{E}[g_t\mid s_{t-1}]=\sum_{a\in\mathcal{V}}\pi_\theta(a\mid s_{t-1})\nabla_\theta\log\pi_\theta(a\mid s_{t-1})=\nabla_\theta\sum_{a\in\mathcal{V}}\pi_\theta(a\mid s_{t-1})=0$.
Below we write $\sum_{t=1}^{T}=\sum_{t=1}^{H}\mathbf{1}\{t\le T\}$ and apply such identities term by term.

\subsection{What the Score Measures}\label{app:theory_lemma}

To interpret the raw score $m_t$, i.e., the mean similarity to other correct solutions minus that to other incorrect ones, define the average similarity and the similarity-weighted success rate
\[
\bar k(s)=\mathbb{E}_{y'}\big[k(s,y')\big],\qquad
P_k(s)=\frac{\mathbb{E}_{y'}\big[k(s,y')\,r(y')\big]}{\bar k(s)},\qquad y'\sim\pi_\theta(\cdot\mid q),
\]
assuming $\bar k(s)>0$. Since $k\ge 0$, $P_k(s)\in[0,1]$ is a weighted average of success indicators, in which each solution is weighted by its similarity to $s$.

\paragraph{Lemma 1 (What the Score Measures).}
\textit{Let $0<p_q<1$. For every prefix $s$ with $t\ge 1$ tokens of the $i$-th solution and $\bar k(s)>0$,}
\[
\mathbb{E}\big[m_t\mid s_t=s,\ E_i\big]=\frac{\bar k(s)}{p_q(1-p_q)}\big(P_k(s)-p_q\big).
\]

\noindent\textit{Proof.}
Let $\mathbf{r}_{-i}=\big(r(y^{(j)})\big)_{j\ne i}$ be the labels of the other solutions; $E_i$, $\mathcal{C}_i$ and $\mathcal{W}_i$ are functions of $\mathbf{r}_{-i}$. Since the $G$ solutions are independent, $(y^{(j)})_{j\ne i}$ is independent of $y^{(i)}$. Moreover, conditional on $\mathbf{r}_{-i}$, each $y^{(j)}$, $j\ne i$, follows $\pi_\theta(\cdot\mid q)$ conditioned on $r(y')=r(y^{(j)})$. Let $\mu_b(s)=\mathbb{E}_{y'}\big[k(s,y')\mid r(y')=b\big]$ for $b\in\{0,1\}$. For $j\ne i$, the similarity $k_t(y^{(j)})$ depends on $y^{(i)}$ and $y^{(j)}$; averaging first over the continuation of $s$ (which is independent of $y^{(j)}$ and $\mathbf{r}_{-i}$) and then over $y^{(j)}$ gives
\[
\mathbb{E}\big[k_t(y^{(j)})\mid s_t=s,\ \mathbf{r}_{-i}\big]
=\mathbb{E}_{y'}\big[k(s,y')\mid r(y')=r(y^{(j)})\big]=\mu_{r(y^{(j)})}(s).
\]
On $E_i$, linearity then gives
\[
\mathbb{E}\big[m_t\mid s_t=s,\ \mathbf{r}_{-i}\big]
=\frac{1}{|\mathcal{C}_i|}\sum_{j\in\mathcal{C}_i}\mu_1(s)-\frac{1}{|\mathcal{W}_i|}\sum_{j\in\mathcal{W}_i}\mu_0(s)
=\mu_1(s)-\mu_0(s).
\]
This value does not depend on $\mathbf{r}_{-i}$, so averaging over $\mathbf{r}_{-i}$ on $E_i$ gives $\mathbb{E}[m_t\mid s_t=s,E_i]=\mu_1(s)-\mu_0(s)$. Here $\Pr(E_i)=1-p_q^{G-1}-(1-p_q)^{G-1}>0$, because $G\ge 3$ and $0<p_q<1$. Finally, $\mathbb{E}_{y'}[k\,r]=\bar k\,P_k$ and $\mathbb{E}_{y'}[k\,(1-r)]=\bar k\,(1-P_k)$, so, with all quantities evaluated at $s$,
\[
\mu_1-\mu_0=\frac{\bar k\,P_k}{p_q}-\frac{\bar k\,(1-P_k)}{1-p_q}
=\bar k\,\frac{P_k(1-p_q)-p_q(1-P_k)}{p_q(1-p_q)}
=\frac{\bar k}{p_q(1-p_q)}\big(P_k-p_q\big).
\]
\hfill$\square$

\textbf{Intuition and Remark on Lemma~1.} Lemma~1 shows that the expected raw score is positive when similarity-weighted success exceeds the question's average. However, $P_k$ weights other solutions by similarity, whereas $P^\pi$ conditions on the actual prefix. Using the former as a proxy requires the representation assumption $P_k\approx P^\pi$; reconstruction and sparsity alone do not establish it. This explains the prefix score, but not yet how it credits the token just added. We therefore examine its role in the update using Proposition~1 below.

\subsection{Credit in the Expected Update}\label{app:theory_credit}

For a prefix $s$ with $t\ge 1$ tokens, define $f(s)=\mathbb{E}\big[\hat S_t\mid s_t=s\big]$. The expectation is over everything else that $\hat S_t$ depends on: the other solutions of the group, the rest of the batch through $\mu_B$ and $\sigma_B$, and the continuation of $y$ (for example through $\rho$ and the BatchTopK threshold). Let $f(s_0)$ be an arbitrary fixed value.

\paragraph{Proposition 1 (Credit in the Expected Update).}
\textit{The expected SAE term satisfies}
\[
\mathbb{E}\Big[\sum_{t=1}^{T}\hat S_t\,g_t\Big]
=\mathbb{E}\Big[\sum_{t=1}^{T}f(s_t)\,g_t\Big]
=\mathbb{E}\Big[\sum_{t=1}^{T}\big(f(s_t)-f(s_{t-1})\big)\,g_t\Big].
\]

\noindent\textit{Proof.}
For each $t$, the factors $\mathbf{1}\{t\le T\}$ and $g_t$ are determined by $s_t$, so by the tower property
\[
\mathbb{E}\big[\mathbf{1}\{t\le T\}\,\hat S_t\,g_t\big]
=\mathbb{E}\big[\mathbf{1}\{t\le T\}\,\mathbb{E}[\hat S_t\mid s_t]\,g_t\big]
=\mathbb{E}\big[\mathbf{1}\{t\le T\}\,f(s_t)\,g_t\big].
\]
Summing over $t$ gives the first equality. For the second, Eq.~\eqref{eq:app_theory_baseline} with $h=f$ gives $\mathbb{E}\big[\mathbf{1}\{t\le T\}f(s_{t-1})g_t\big]=0$ for every $t$; for $t=1$, $f(s_0)$ is a fixed value. \hfill$\square$

\textbf{Intuition and Remark on Proposition~1.} In other words, in expectation SAEPO credits each token by how much it raises or lowers the expected score, i.e., $f(s_t)-f(s_{t-1})$, just as the exact policy gradient of the success rate credits each token by how much it raises or lowers the success probability $P^\pi$ (Proposition~3). This is an equivalence of expected updates, not of individual sampled scores: a sampled $\hat S_t$ can differ from $f(s_t)-f(s_{t-1})$, but the two give the same expected update. Lemma~1 links this credit to $P_k$; Corollary~1 below makes that connection explicit when the reliability weight and normalization statistics are held fixed.

\paragraph{Corollary 1 (Score Differences and $P_k$).}
\textit{Let $0<p_q<1$, replace $\rho$, $\mu_B$ and $\sigma_B$ by deterministic constants with $\rho>0$ and $\sigma_B\ge 0$, and let $\kappa_q=\Pr(E_i)=1-p_q^{G-1}-(1-p_q)^{G-1}$. Then, for every prefix $s$ with $t\ge 1$ tokens and $\bar k(s)>0$,}
\[
f(s)=\frac{\kappa_q}{\sigma_B+\epsilon_{\mathrm{score}}}\Big(\frac{\rho\,\bar k(s)}{p_q(1-p_q)}\big(P_k(s)-p_q\big)-\mu_B\Big).
\]
\textit{For a transition with $t\ge 2$ and $\bar k(s_t)=\bar k(s_{t-1})=\bar k$, it follows that}
\[
f(s_t)-f(s_{t-1})=\lambda\,\big(P_k(s_t)-P_k(s_{t-1})\big),\qquad
\lambda=\frac{\kappa_q\,\rho\,\bar k}{(\sigma_B+\epsilon_{\mathrm{score}})\,p_q(1-p_q)}>0,
\]
\textit{where $\lambda$ depends on the transition only through $\bar k$.}

\noindent\textit{Proof.}
With these constants, $\hat S_t=\mathbf{1}_{E_i}\,(\rho\,m_t-\mu_B)/(\sigma_B+\epsilon_{\mathrm{score}})$. Since $E_i$ depends only on the other solutions, $\Pr(E_i\mid s_t=s)=\kappa_q$, and therefore
$f(s)=\kappa_q\big(\rho\,\mathbb{E}[m_t\mid s_t=s,E_i]-\mu_B\big)/(\sigma_B+\epsilon_{\mathrm{score}})$. Substituting Lemma~1 gives the first formula. In the difference, the terms that do not depend on the prefix cancel, which gives the second formula. \hfill$\square$

Under these conditions, Proposition~1 shows that, for such transitions, the expected SAE update credits token $y_t$ by $\lambda\big(P_k(s_t)-P_k(s_{t-1})\big)$, the change in similarity-weighted success that the token produces. In the implementation, $\rho$ counts the scored solution's own label, and $\mu_B$ and $\sigma_B$ are batch statistics, so Corollary~1 is an idealization. Whether changes in $P_k$ track changes in $P^\pi$ depends on the representation assumption $P_k\approx P^\pi$.

\subsection{The Surrogate Objective of the SAE Term}\label{app:theory_surrogate}

Fix the current parameters $\theta_0$, write $f=f_{\theta_0}$, and let the prefixes be visited by $\pi_{\theta_0}$. For a prefix $s$ and a token $a$, let $(s,a)$ denote $s$ followed by $a$. Define the surrogate
\[
\mathcal{L}_{\theta_0}(\theta)=\mathbb{E}_q\,\mathbb{E}_{y\sim\pi_{\theta_0}(\cdot\mid q)}\Big[\sum_{t=1}^{T}\ \sum_{a\in\mathcal{V}}\pi_\theta(a\mid s_{t-1})\,f\big((s_{t-1},a)\big)\Big],
\]
the expected score of the next token when this token is drawn from $\pi_\theta$, while the visited prefixes and the score function stay at $\theta_0$.

\paragraph{Proposition 2 (The Update and Its Objective).}
\textit{The SAE term is $\alpha$ times the gradient of the surrogate at $\theta_0$:}
\[
d(\theta_0)=\alpha\,\nabla_\theta\mathcal{L}_{\theta_0}(\theta)\big|_{\theta=\theta_0}.
\]

\noindent\textit{Proof.}
For a fixed prefix $s$, since $\mathcal{V}$ is finite,
\begin{align*}
\nabla_\theta\sum_{a\in\mathcal{V}}\pi_\theta(a\mid s)\,f\big((s,a)\big)\Big|_{\theta=\theta_0}
&=\sum_{a\in\mathcal{V}}\pi_{\theta_0}(a\mid s)\,\nabla_\theta\log\pi_\theta(a\mid s)\big|_{\theta=\theta_0}\,f\big((s,a)\big)\\
&=\mathbb{E}\big[f(s_t)\,g_t\mid s_{t-1}=s\big],
\end{align*}
where $y_t\sim\pi_{\theta_0}(\cdot\mid s)$. Multiplying by $\mathbf{1}\{t\le T\}$, which is determined by $s_{t-1}$, taking the expectation over the prefixes visited by $\pi_{\theta_0}$, and summing over $t$ gives $\nabla_\theta\mathcal{L}_{\theta_0}(\theta_0)=\mathbb{E}\big[\sum_{t}f(s_t)\,g_t\big]$. By the first equality of Proposition~1, this equals $\mathbb{E}\big[\sum_t\hat S_t\,g_t\big]=d(\theta_0)/\alpha$. \hfill$\square$

\textbf{Intuition and Remark on Proposition~2.} As in PPO, the surrogate is rebuilt after each update, holding $f$ and the prefix distribution fixed for that update. Intuitively, Proposition~2 implies that at each prefix, the SAE term moves probability toward next tokens whose new prefix has an above-average expected score, i.e., roughly, is more similar to the group's correct and less to its incorrect solutions. However, raising this score alone need not raise the success rate. To address this, Proposition~3 below further connects the SAE term to the success rate.

\subsection{Relation to the Expected Success Rate}\label{app:theory_return}

The original objective is $J(\theta)=\mathbb{E}_q\big[p_q(\theta)\big]$. The exact policy gradient of the success rate credits token $y_t$ with the change in success probability it causes, whereas the SAE term credits it with the change in expected score, $f(s_t)-f(s_{t-1})$ (Proposition~1). To compare the two, fix a scale $c_q>0$ and an offset $b_q$ per question and define the remainder $e$, the part of the expected score not explained by success probability, by
\[
f(s)=c_q\,P^\pi(s)+b_q+e(s)
\]
for every prefix $s$ of question $q$. This identity defines $e$; it does not assume that $e$ is small. The remainder absorbs all discrepancies, including those from representation similarity, prefix-dependent $\bar k$, normalization and reliability weighting. Let $\bar e(s_{t-1})=\mathbb{E}\big[e(s_t)\mid s_{t-1}\big]$ with $y_t\sim\pi_\theta(\cdot\mid s_{t-1})$, let $g_c=\mathbb{E}_q\big[c_q\nabla_\theta p_q\big]$, and let $J_c(\theta')=\mathbb{E}_q\big[c_q\,p_q(\theta')\big]$ with the weights $c_q$ held fixed, so that $\nabla_\theta J_c(\theta)=g_c$.

\paragraph{Proposition 3 (Relation to the Expected Return).}
\textit{(i) For every question $q$, with $y\sim\pi_\theta(\cdot\mid q)$,}
\[
\nabla_\theta p_q=\mathbb{E}\Big[\sum_{t=1}^{T}\big(P^\pi(s_t)-P^\pi(s_{t-1})\big)\,g_t\ \Big|\ q\Big].
\]
\textit{(ii) The SAE term decomposes as}
\[
d(\theta)=\alpha\big(g_c+R\big),\qquad
R=\mathbb{E}\Big[\sum_{t=1}^{T}\big(e(s_t)-\bar e(s_{t-1})\big)\,g_t\Big].
\]
\textit{(iii) If $\lVert g_t\rVert\le L$ and $|e(s_t)-\bar e(s_{t-1})|\le\delta$ for all reachable prefixes, then $\lVert R\rVert\le HL\delta$.}

\noindent\textit{(iv) $g_c^\top d(\theta)\ge\alpha\lVert g_c\rVert\big(\lVert g_c\rVert-\lVert R\rVert\big)$. Hence, if $\lVert R\rVert<\lVert g_c\rVert$, a sufficiently small step along $d(\theta)$ increases $J_c$.}

\noindent\textit{Proof.}
(i) The probability of a solution is $\prod_{t=1}^{T}\pi_\theta(y_t\mid s_{t-1})$ and the set of solutions is finite, so $\nabla_\theta p_q=\mathbb{E}\big[r(y)\sum_{t=1}^{T}g_t\mid q\big]$. For each $t$, the factors $\mathbf{1}\{t\le T\}$ and $g_t$ are determined by $s_t$, and $\mathbb{E}[r(y)\mid s_t]=P^\pi(s_t)$, so $\mathbb{E}\big[\mathbf{1}\{t\le T\}\,r(y)\,g_t\mid q\big]=\mathbb{E}\big[\mathbf{1}\{t\le T\}\,P^\pi(s_t)\,g_t\mid q\big]$. Subtracting $\mathbb{E}\big[\mathbf{1}\{t\le T\}\,P^\pi(s_{t-1})\,g_t\mid q\big]=0$ (Eq.~\eqref{eq:app_theory_baseline}) and summing over $t$ gives (i).

(ii) By Proposition~1, $d(\theta)=\alpha\,\mathbb{E}\big[\sum_t f(s_t)\,g_t\big]$. Conditional on its question $q$, the solution $y$ is drawn from $\pi_\theta(\cdot\mid q)$. Substitute $f=c_qP^\pi+b_q+e$ and condition on $q$. The first term gives $c_q\,\mathbb{E}\big[\sum_t P^\pi(s_t)\,g_t\mid q\big]=c_q\nabla_\theta p_q$, by (i) and Eq.~\eqref{eq:app_theory_baseline}. The second term vanishes by Eq.~\eqref{eq:app_theory_baseline}. The third term equals $\mathbb{E}\big[\sum_t\big(e(s_t)-\bar e(s_{t-1})\big)g_t\mid q\big]$, by Eq.~\eqref{eq:app_theory_baseline} with $h=\bar e$. Averaging over questions gives (ii).

(iii) By the triangle inequality,
$\lVert R\rVert\le\mathbb{E}\big[\sum_{t=1}^{T}|e(s_t)-\bar e(s_{t-1})|\,\lVert g_t\rVert\big]\le\mathbb{E}[T]\,\delta L\le HL\delta$.

(iv) By the Cauchy--Schwarz inequality, $g_c^\top d=\alpha\big(\lVert g_c\rVert^2+g_c^\top R\big)\ge\alpha\big(\lVert g_c\rVert^2-\lVert g_c\rVert\,\lVert R\rVert\big)$. If $\lVert R\rVert<\lVert g_c\rVert$, then $g_c\ne 0$ and $g_c^\top d>0$. Since $J_c$ is differentiable with gradient $g_c$, $J_c(\theta+\eta d)=J_c(\theta)+\eta\,g_c^\top d+o(\eta)>J_c(\theta)$ for all sufficiently small $\eta>0$. \hfill$\square$

\paragraph{Remarks.}
(a) \emph{Which Errors Matter.} Let $\bar f(s_{t-1})=\mathbb{E}[f(s_t)\mid s_{t-1}]$. Since $\mathbb{E}[P^\pi(s_t)\mid s_{t-1}]=P^\pi(s_{t-1})$,
\[
e(s_t)-\bar e(s_{t-1})=\big(f(s_t)-\bar f(s_{t-1})\big)-c_q\big(P^\pi(s_t)-P^\pi(s_{t-1})\big).
\]
Hence $R$ depends only on how $f$ and $c_qP^\pi$ compare alternative next tokens at the same prefix. An error that is shared by all next tokens at a prefix, and a per-question offset, cancel.

(b) \emph{Comparison with GRPO: Question Weighting.} The SAE term changes GRPO's policy update. We make this change explicit by comparing both updates with the gradient of the expected return $J$, i.e., the average success rate that both methods aim to increase. Apart from $R$, the SAE term follows the gradient of the weighted success rate $J_c$ rather than of $J$: the weights $c_q>0$ change how much each question counts, and $J_c$ is a positive scalar multiple of $J$ if all $c_q$ are equal. GRPO's division by the group's reward standard deviation also reweights questions. Consider the per-solution summed update in an idealized analysis with binary rewards, in which the group mean and standard deviation of the rewards are replaced by their population values $p_q$ and $\sqrt{p_q(1-p_q)}$ (the large-group limit). For a question with $0<p_q<1$, $A=(r(y)-p_q)/\sqrt{p_q(1-p_q)}$, and using $\mathbb{E}\big[r(y)\sum_t g_t\mid q\big]=\nabla_\theta p_q$ and $\mathbb{E}\big[\sum_t g_t\mid q\big]=0$,
\[
\mathbb{E}\Big[\sum_{t=1}^{T}A\,g_t\ \Big|\ q\Big]
=\frac{\nabla_\theta p_q}{\sqrt{p_q(1-p_q)}}
=\nabla_\theta\,2\arcsin\sqrt{p_q},
\]
because $\frac{d}{dp}\,2\arcsin\sqrt{p}=1/\sqrt{p(1-p)}$. If $p_q\in\{0,1\}$, all solutions of $q$ have the same label; we define $A=0$, and $p_q$ does not depend on $\theta$, so these questions contribute zero gradient. Averaging over questions, this idealized GRPO outcome term is the gradient of $\mathbb{E}_q\big[2\arcsin\sqrt{p_q}\big]$, which weights each question with $0<p_q<1$ by $1/\sqrt{p_q(1-p_q)}$. Writing $w_q=1/\sqrt{p_q(1-p_q)}$ for these questions and $w_q=0$ at the boundaries, Proposition~3 gives the combined per-solution summed outcome and SAE update under the same on-policy idealization, excluding the KL term:
\[
\mathbb{E}\Big[\sum_{t=1}^{T}(A+\alpha\hat S_t)\,g_t\Big]
=\mathbb{E}_q\big[(w_q+\alpha c_q)\nabla_\theta p_q\big]+\alpha R.
\]
Relative to GRPO, SAEPO therefore changes the update in only two ways: it changes the weight of each question from $w_q$ to $w_q+\alpha c_q$, which stays positive, and it adds the error $\alpha R$ discussed above.

(c) \emph{Connection to Empirical Results.} Note that here we do not claim that $\delta$ is small everywhere. The weight $\alpha$ scales both $g_c$ and $R$ relative to the outcome term. We also provide empirical results from controlled edits and sampled continuations in Section~\ref{sec:results} to support the theoretical analysis. Specifically, the experiments examine the two-hop connection: \emph{SAEPO's token-level semantic score} $\rightarrow$ \emph{correctness of a local step} $\rightarrow$ \emph{prefix-continuation success rates (final accuracy)}. Table~\ref{tab:controlled_step_edits} evaluates the first connection, and Table~\ref{tab:controlled_continuations} evaluates the second. Together, these theoretical and empirical results connect SAEPO's token-level scores to local reasoning quality and continuation success.

\section{Training Details}
\label{app:training_details}

\subsection{Training Data}
\label{app:training_data}

For math, RL runs use the \texttt{DeepScaleR-\allowbreak{}Preview-\allowbreak{}Dataset} \citep{deepscaler2025}, loaded from \nolinkurl{agentica-org/DeepScaleR-Preview-Dataset}. Its training split contains \(40{,}315\) math problems from competition-style sources, including AIME, AMC, Omni-MATH, and STILL. We do not apply subject filtering, a difficulty curriculum, or additional subsampling. Matched methods use the same training pool and data order.

For code generation, we use the TACO, PrimeIntellect SYNTHETIC-1, and LiveCodeBench-v5 training splits of the \texttt{DeepCoder-\allowbreak{}Preview-\allowbreak{}Dataset} \citep{deepcoder2025}. These splits provide executable unit tests, and matched methods use the same problems.

\paragraph{Disjoint from evaluation.}
The DeepScaleR training pool is disjoint from the six-benchmark math evaluation suite. Its internal AIME content is restricted to historical (pre-2024) competitions, so it is also disjoint from AIME25.

\subsection{Prompt and Generation Settings}
\label{app:prompt_generation}

For math, training rollouts and downstream evaluation use the same
\textsc{minerva\_math} prompt template. The demonstrations are three
worked-out math problems plus the target problem, in the format
\begin{quote}\small
\texttt{Problem:\textbackslash n\allowbreak{}\{example\_1\}\textbackslash n\allowbreak{}\textbackslash n\allowbreak{}Solution:\textbackslash n\allowbreak{}\{worked\_solution\_1\}\textbackslash n\allowbreak{}Final Answer: ...\textbackslash n\allowbreak{}\textbackslash n\allowbreak{}}\,(\(\times 3\))\,\texttt{Problem:\textbackslash n\allowbreak{}\{problem\}\textbackslash n\allowbreak{}\textbackslash n\allowbreak{}Solution:\textbackslash n\allowbreak{}}
\end{quote}
where each demonstration solution terminates in a
\texttt{\textbackslash boxed\{...\}} expression so the model copies that
surface form. The string \texttt{Problem:} is registered as a vLLM stop
token so a generation cannot leak into a hallucinated next problem; the
model EOS token \texttt{<|endoftext|>} is the other stop signal.

All runs sample \(G=8\) rollouts per question at temperature \(T=1.0\); evaluation uses greedy decoding. The maximum generation length is \(8192\) tokens for math and \(4096\) for code. For the Qwen math runs, we use vLLM 0.16 with top-\(p=1.0\), maximum prompt length \(1280\), and no length penalty. Weights are synchronized from the HuggingFace policy to vLLM every \(10\) training steps via CuMemAllocator wake/sleep, with \texttt{gpu\_memory\_utilization} set to \(0.50\) for 1.7B and \(0.85\) for 4B.

\subsection{Reward Construction}
\label{app:reward}

For math, the scalar outcome reward is \(R_{\mathrm{out}}=R_{\mathrm{acc}}+R_{\mathrm{fmt}}\), where \(R_{\mathrm{acc}}\in\{0,1\}\) records final-answer correctness and \(R_{\mathrm{fmt}}\in\{0,0.1\}\) rewards a boxed answer. We extract the final boxed answer and grade it with \texttt{math\_verify}. For code, \(R_{\mathrm{out}}\in\{0,1\}\) is one only when every unit test for the problem passes.

The GRPO/GSPO outcome advantage \(A^{\mathrm{out}}_{qg}\) is computed from \(R_{\mathrm{out}}\). For \modelname{}, the correct/incorrect split uses \(R_{\mathrm{acc}}\) for math and test-suite success for code; a math rollout with only the format bonus remains incorrect. After group normalization of the outcome reward and batch standardization of the semantic score, only the advantage changes, as in~\eqnref{eq:final-adv}.

\subsection{Optimization Hyperparameters}
\label{app:optimization}

Within each matched vanilla/\modelname{} pair, the verifier reward, policy-ratio construction, clipping rule, KL regularizer, sampling configuration, and optimizer hyperparameters are identical. \modelname{} replaces the broadcast rollout-level advantage \(A^{\mathrm{out}}_{qg}\) with the token-level advantage \(A_{qg,t}\) from~\eqnref{eq:final-adv}.

All runs use full-parameter fine-tuning, without LoRA or adapters. We
use \texttt{torch.optim.AdamW} with \(\beta_1=0.9\), \(\beta_2=0.999\),
weight decay \(0\), and numerical constant \(\epsilon=10^{-8}\). The
learning rate is \(10^{-5}\) on a constant schedule (no warmup, no
decay). We use KL coefficient \(\beta=0.02\) against a frozen
reference policy \(\pi_{\mathrm{ref}}\) (the SFT initialization),
bf16 mixed precision, gradient checkpointing, micro-batch size \(4\),
and \(4\) gradient-accumulation steps. Randomness is controlled with
seed \(42\) for Python, NumPy, PyTorch, vLLM, and rollout sampling.

The additional random-seed comparison in Appendix~\ref{app:multiple_seeds} also includes seeds 43 and 44.

The Qwen3-4B and OLMo-3-7B runs use \(8\) questions per batch, corresponding to
  \(64\) rollouts per training step. The Qwen3-1.7B runs use \(32\)
  questions per batch, corresponding to \(256\) rollouts per training
  step. The 4B and 7B runs are trained for \(1000\) steps and the 1.7B runs
  for \(250\) steps. Both settings therefore consume the same budget of
  \(64{,}000\) rollouts, so results across model scales are compared at
  matched sampling cost.
All main results are reported at the last checkpoint
rather than selecting the best checkpoint independently for each method.

\paragraph{Single-GPU compute.} Each math run occupies a single NVIDIA
RTX PRO 6000 Blackwell ($\sim$\(97\) GB HBM); vLLM, the HF policy,
\(\pi_{\mathrm{ref}}\), and the SAE all share that GPU through
CuMemAllocator wake/sleep. End-to-end wall-clock is approximately \(80\) hours for a 7B run,
\(36\) hours for a 4B run and \(14\) hours for a 1.7B run. Training,
evaluation, and SAE encoding are all single-GPU and require no
inter-node communication.

\subsection{Computational Cost and Wall-clock Efficiency}
\label{app:compute_efficiency}

We ran additional experiments to document the time and peak GPU memory consumption of all computation steps in \modelname{} and vanilla GRPO/GSPO RL training on \texttt{Qwen3-\allowbreak{}1.7B-\allowbreak{}Base}. Results are summarized in Table~\ref{tab:rebuttal_compute}.

\begin{table}[tbp]
\centering
\small
\setlength{\tabcolsep}{4pt}
\caption{Time and memory for \texttt{Qwen3-\allowbreak{}1.7B-\allowbreak{}Base}. Dedicated SAE feature extraction and semantic alignment add 1.11 s per GRPO step and 0.97 s per GSPO step, while peak GPU memory increases by 1.1 GB. The 250-step wall-clock times cover RL training.}
\label{tab:rebuttal_compute}
\begin{tabular}{lcccc}
\toprule
Component / measurement & Vanilla GRPO & \modelname{} (GRPO) & Vanilla GSPO & \modelname{} (GSPO) \\
\midrule
Rollout generation (per step) & 33.1 s & 37.8 s & 32.9 s & 34.5 s \\
Reward grading (per step) & 3.5 s & 2.7 s & 3.8 s & 2.7 s \\
Old/reference log-probs (per step) & 12.3 s & 13.6 s & 12.3 s & 13.1 s \\
Policy update (per step) & 32.4 s & 36.6 s & 32.7 s & 34.7 s \\
SAE feature extraction (per step) & -- & 0.77 s & -- & 0.68 s \\
Semantic alignment (per step) & -- & 0.34 s & -- & 0.29 s \\
\midrule
\textbf{Total per step} & \textbf{81.4 s} & \textbf{92.0 s} & \textbf{81.8 s} & \textbf{86.0 s} \\
\midrule
Wall-clock, 250 RL steps & 5.66 h & 6.39 h & 5.68 h & 5.97 h \\
Peak GPU memory & 17.3 GB & 18.4 GB & 17.3 GB & 18.4 GB \\
\bottomrule
\end{tabular}%

\end{table}

We can see that during training, the extra time introduced by the SAE consists of token feature extraction and cross-rollout semantic alignment: 1.11 s per step for GRPO, i.e., 1.4\% of a vanilla GRPO step (81.4 s), and 0.97 s per step for GSPO, i.e., 1.2\% of a vanilla GSPO step (81.8 s). The rest of the difference in total time comes from \modelname{} policies generating longer responses. Therefore, the direct computational overhead of SAE feature extraction and semantic alignment is minimal.

Note that SAE pretraining is a one-time cost: the SAE is trained once per backbone and reused across runs. Our SAE used by \texttt{Qwen3-\allowbreak{}1.7B-\allowbreak{}Base} remains unchanged for both math and code and takes 8.1 GPU-hours to pretrain. On our single-GPU setup, this pretraining plus one 250-step \modelname{} RL run takes approximately 14 hours in total. Next, we compare \modelname{} with vanilla GRPO/GSPO under matched RL wall-clock budgets and report the training-set accuracies across methods in Table~\ref{tab:rebuttal_wallclock}, with the pretrained SAE reused.

\begin{table}[tbp]
\centering
\small
\setlength{\tabcolsep}{4pt}
\caption{Training accuracy (\%) under matched RL wall-clock budgets on \texttt{Qwen3-\allowbreak{}1.7B-\allowbreak{}Base}. \modelname{} improves accuracy over the corresponding vanilla optimizer at every reported time point. These budgets begin with RL training and reuse the pretrained SAE.}
\label{tab:rebuttal_wallclock}
\begin{tabular}{lcccc}
\toprule
RL wall-clock & Vanilla GRPO & \modelname{} (GRPO) & Vanilla GSPO & \modelname{} (GSPO) \\
\midrule
1 h & 12.9 & \textbf{15.8} & 11.8 & \textbf{13.3} \\
2 h & 16.9 & \textbf{25.8} & 14.6 & \textbf{21.2} \\
3 h & 20.0 & \textbf{28.1} & 17.3 & \textbf{28.0} \\
4 h & 20.0 & \textbf{28.8} & 20.1 & \textbf{24.9} \\
5 h & 20.7 & \textbf{28.2} & 21.6 & \textbf{27.4} \\
5.66 h & 22.2 & \textbf{28.7} & 20.7 & \textbf{25.5} \\
\bottomrule
\end{tabular}%
\end{table}

These results show that \modelname{} reaches the respective vanilla optimizers' final training accuracies after 1.35 hours (GRPO) and 1.75 hours (GSPO), while vanilla GRPO and vanilla GSPO need 5.66 hours and 5.68 hours, respectively. This highlights \modelname{}'s advantage in accelerating convergence in RL training.

\subsection{Sparse Autoencoder Details}
\label{app:sae_training}

For each backbone, we train a BatchTopK SAE \citep{bussmann2024batchtopk} on target-layer hidden activations before RL. During RL, the frozen encoder provides \(\phi_{qg,t}=f_{\mathrm{enc}}(H_{qg,t})\) for cross-rollout semantic alignment in~\eqnref{eq:soft-align}; the SAE is not used at inference time.

For \texttt{Qwen3-\allowbreak{}4B-\allowbreak{}Base}, the SAE uses layer \(27\), hidden dimension \(2560\), a \(32\times\) expansion, and \(81{,}920\) latent features. For \texttt{Qwen3-\allowbreak{}1.7B-\allowbreak{}Base}, it uses layer \(21\), hidden dimension \(2048\), a \(32\times\) expansion, and \(65{,}536\) latent features. The target average sparsity is \(k=16\) for 4B and \(k=8\) for 1.7B. The \texttt{OLMo-\allowbreak{}3-\allowbreak{}7B-\allowbreak{}Base} SAE uses layer \(21\), a \(32\times\) expansion, and target average sparsity \(k=16\). The Qwen SAEs are trained on FineWeb-Edu hidden activations dumped from the corresponding base models using the EleutherAI \texttt{sparsify} library, with Adam and learning rate \(10^{-4}\).

The SAE parameters remain fixed throughout policy optimization, and no SAE forward pass is required at inference time. We set \(\alpha=2\) and \(\tau=0.05\) for all three backbones and both tasks.

\section{Evaluation Details}
\label{app:evaluation_details}

We evaluate accuracy with greedy decoding on six math benchmarks:
MATH-500 \citep{lightman2023letsverifystepstep},
GSM8K \citep{cobbe2021trainingverifierssolvemath},
AIME25 \citep{aime25},
AMC \citep{numina_math_datasets},
Minerva Math \citep{lewkowycz2022solvingquantitativereasoningproblems},
and OlympiadBench \citep{he2024olympiadbenchchallengingbenchmarkpromoting}.
The number of problems per benchmark is
\((500,\allowbreak 1319,\allowbreak 30,\allowbreak 83,\allowbreak 272,\allowbreak 674)\), for a total of \(2878\) problems. We
report the unweighted average across these six benchmarks. AIME25 has
only \(30\) problems, so one problem changes its accuracy by \(3.3\)
percentage points. For math, answers are extracted as the last boxed
expression and graded by \texttt{math\_verify}.

For code generation, we evaluate on LiveCodeBench-v5 \citep{jain2024livecodebench,deepcoder2025} and LeetCode Dataset v0.3.1 \citep{xia2025leetcodedatasettemporaldatasetrobust} (228 held-out problems). A problem counts as solved only if all its unit tests pass. The reported code average weights the two benchmarks equally.

Unless otherwise stated, comparisons use endpoint checkpoints under matched training budgets. This avoids selecting each method's best checkpoint independently.

\section{Interpretability Analysis Details}
\label{app:interpretability_details}

\subsection{Feature-Level Heatmap Construction}
\label{app:feature_heatmap}

\paragraph{Stage 1: Rollout pool.}
We draw a stratified subsample of \(200\) problems from the MATH-500 test split. The seven subject labels are Algebra, Geometry, Number Theory, Counting~\&~Probability, Prealgebra, Intermediate Algebra, and Precalculus. Sampling uses \texttt{numpy.random.default\_rng} with seed \(42\), with approximately \(200/7\) problems per subject. For each question, we generate \(G=8\) rollouts with \texttt{Qwen3-\allowbreak{}4B-\allowbreak{}Base} at \(T=1.0\), \(\textit{max\_new\_tokens}=512\), and the math prompt described in~\appref{app:prompt_generation}. We grade each rollout with \texttt{math\_verify} and keep groups with at least two correct and two incorrect rollouts, so every retained rollout has both comparison sets after self-exclusion.

\paragraph{Stage 2: SAE encoding.}
For every retained rollout, we capture each generated token's layer-\(27\) hidden state and encode it with the frozen Qwen3-4B SAE. We store the \((\texttt{top\_idx},\texttt{top\_acts})\) pairs, with \(k=16\) active features per token on average and \(81{,}920\) latent features in total.

\paragraph{Stage 3: Cross-rollout semantic alignment.}
For each token, we compute soft alignment against all generated positions in every other rollout of the same question, then average those alignment scores over correct and incorrect rollouts separately. The resulting raw semantic score \(S_{qg,t}\) follows~\eqnref{eq:soft-align}--\eqnref{eq:raw-score}. The token's own rollout is excluded, and rollouts need not align by generation position.

\paragraph{Stage 4: Feature-level aggregation.}
For each subject and SAE feature, we average \(S_{qg,t}\) over tokens where that feature is active. We rank features by this mean score and display the five highest and five lowest for each subject in the figure. Each row lists the four token strings with highest mean activation for that feature; token strings must occur at least twice, and displayed features must fire at least \(30\) times. Cell color is normalized within each feature row, showing relative activation among its tokens.

\begin{table}[t]
\centering
\small
\setlength{\tabcolsep}{4pt}
\caption{Ablation study of SAE hook layer selection for
\texttt{Qwen3-\allowbreak{}1.7B-\allowbreak{}Base} with GRPO + SAEPO. We report accuracy (\%) on six
math benchmarks and their unweighted average. Boldface marks the best
value in each column.}
\label{tab:layer}
\begin{tabular}{l cccccc c}
\toprule
Layer & MATH-500 & GSM8K & AIME25 & AMC & Minerva & Olympiad & Avg. \\
\midrule
Layer 3  & \textbf{66.0} & 74.5 & \textbf{3.3} & 25.3 & 28.7 & 23.1 & 36.82 \\
Layer 9  & 64.4 & 75.6 & \textbf{3.3} & 24.1 & 29.4 & \textbf{23.7} & 36.75 \\
Layer 15 & 64.8 & 70.2 & \textbf{3.3} & 25.3 & \textbf{29.8} & 22.1 & 35.92 \\
\textbf{Layer 21} & 65.6 & \textbf{76.3} & \textbf{3.3} & \textbf{34.9} & \textbf{29.8} & 22.1 & \textbf{38.68} \\
Layer 27 & 64.4 & 75.9 & 0.0 & 33.7 & 27.6 & 22.4 & 37.33 \\
\bottomrule
\end{tabular}
\end{table}

\begin{table}[t]
\centering
\small
\setlength{\tabcolsep}{4pt}
\caption{Ablation of SAE sparsity \(k\) for \texttt{Qwen3-\allowbreak{}1.7B-\allowbreak{}Base} with GRPO + \modelname{}. We report accuracy (\%) on six math benchmarks and their unweighted average.}
\label{tab:sparsity}
\begin{tabular}{l cccccc c}
\toprule
Method & MATH-500 & GSM8K & AIME25 & AMC & Minerva & Olympiad & Avg. \\
\midrule
Vanilla GRPO & 63.2 & 74.9 & \textbf{3.3} & 26.5 & 23.9 & 21.5 & 35.56 \\
\midrule
\modelname{} ($k{=}4$) & 65.6 & 74.7 & \textbf{3.3} & 27.7 & 26.1 & 21.8 & 36.54 \\
\modelname{} ($k{=}8$) & 65.6 & 76.3 & \textbf{3.3} & 34.9 & 29.8 & 22.1 & \textbf{38.68}\\
\modelname{} ($k{=}16$) & 64.6 & \textbf{79.0} & 0.0 & 30.1 & 30.1 & 21.8 & 37.61 \\
\modelname{} ($k{=}32$) & 63.6 & 72.3 & \textbf{3.3} & \textbf{37.3} & 32.4 & 22.1 & 38.50 \\
\modelname{} ($k{=}64$) & \textbf{66.2} & 72.0 & 0.0 & 28.9 & \textbf{33.5} & \textbf{23.4} & 37.34 \\
\bottomrule
\end{tabular}
\end{table}

\section{Ablation Details}
\label{app:ablation_details}

The component ablation appears in Table~\ref{tab:ablation_basis_sign}. Here we give its scoring details and report the SAE sparsity and hook-layer ablations on \texttt{Qwen3-\allowbreak{}1.7B-\allowbreak{}Base} in Tables~\ref{tab:layer} and~\ref{tab:sparsity}.

\paragraph{Raw-Embedding Contrast.}
This ablation replaces the SAE code \(\phi_{qg,t}\) with the raw target-layer hidden state \(H_{qg,t}\), while keeping the cross-rollout soft-alignment and correct-minus-incorrect scoring rules. Its lower average in Table~\ref{tab:ablation_basis_sign} suggests that the sparse feature representation matters for this credit signal.

\paragraph{Reverse \modelname{} Signal.}
This ablation reverses the raw semantic score, \(S^{\mathrm{rev}}_{qg,t}=-S_{qg,t}\), before batch standardization and advantage shaping. Its lower accuracy indicates that the correct-minus-incorrect direction matters.

Table~\ref{tab:sparsity} varies target average sparsity \(k\), with \(k=8\) giving the highest average among the tested values. Table~\ref{tab:layer} varies the SAE hook layer, with layer \(21\) giving the highest reported average.

\subsection{Comparison with Dense Learned Representations}
\label{app:representation_comparison}

In Table~\ref{tab:ablation_basis_sign}, we compare credit assignment using SAE representations with raw hidden states from the language model, and SAE representations show better performance. We run additional experiments to compare SAE, autoencoder (AE), and MLP-based encoders with the same latent dimension of 65,536. For training budgets, we train all three encoders on the policy's own math rollouts: 32,275 trajectories generated by \texttt{Qwen3-\allowbreak{}1.7B-\allowbreak{}Base} from our RL training set, comprising approximately 43.8 million tokens. We train the SAE with reconstruction and sparsity constraints, the AE with reconstruction alone, and the MLP to predict the outcomes of math rollouts using binary cross-entropy through a linear readout. We then freeze all three encoders and run RL training with the \modelname{} pipeline, with only the latent code differing. Table~\ref{tab:rebuttal_representations} reports the accuracies on the six math benchmarks.

\begin{table}[tbp]
\centering
\small
\setlength{\tabcolsep}{4pt}
\caption{Comparison of encoders trained on the same math rollouts with 65,536 latent dimensions. SAE-based credit assignment achieves the highest six-benchmark average, supporting the contribution of sparse semantic representations beyond a general learned transformation.}
\label{tab:rebuttal_representations}
\begin{tabular}{lccccccc}
\toprule
Representation & MATH-500 & GSM8K & AIME25 & AMC & Minerva & OlympiadBench & Avg. \\
\midrule
SAE latent code & \textbf{65.8} & \textbf{78.0} & \textbf{6.7} & \textbf{34.9} & \textbf{33.1} & \textbf{22.3} & \textbf{40.12} \\
MLP latent code & 59.0 & 77.5 & 3.3 & 26.5 & 26.5 & 19.6 & 35.40 \\
AE latent code & 63.8 & 74.1 & 3.3 & \textbf{34.9} & \textbf{33.1} & 21.8 & 38.50 \\
\bottomrule
\end{tabular}%
\end{table}

The comparison in Table~\ref{tab:rebuttal_representations} shows that \modelname{} with sparse semantic decomposition achieves higher average accuracy than the dense AE and MLP representations. SAE representations achieve 40.12\% average accuracy, compared with 38.50\% for the dense AE and 35.40\% for the MLP, gains of 1.62 and 4.72 percentage points, respectively. Moreover, SAE's latent code makes the credit signal interpretable. Each token activates only a few sparse, largely monosemantic latents, so the similarity that determines its credit decomposes into contributions from the concepts it shares with aligned tokens in correct and incorrect rollouts. This lets us identify which concepts are rewarded or penalized and track them across problems and over training. Dense AE or MLP codes spread the same information over many entangled coordinates with no individual meaning; therefore their credit cannot be traced back to human-interpretable concepts in Appendix~\ref{app:interpretability_details}.

\subsection{Similarity Metrics and Alignment}
\label{app:similarity_alignment}

Our \modelname{} is a general framework and can support different similarity metrics and alignment rules. We chose cosine similarity and soft alignment because of their stable performance in our preliminary experiments. We run additional experiments as ablation studies on the choice of similarity metric and alignment rule. Specifically, we first examine whether these choices affect performance on \texttt{Qwen3-\allowbreak{}1.7B-\allowbreak{}Base} with GRPO. We compare cosine similarity with Jaccard similarity, negative $L_2$ distance, and inner product, retaining soft semantic alignment for each metric. We also replace soft semantic alignment with hard alignment, which compares tokens at the same positions across rollouts, while retaining cosine similarity. Unless varied, these and the following ablations use the default $\alpha=2$ and $\tau=0.05$. Table~\ref{tab:rebuttal_similarity_alignment} shows the results.

\begin{table}[tbp]
\centering
\small
\setlength{\tabcolsep}{4pt}
\caption{Similarity and alignment ablations. Cosine similarity with soft alignment achieves the highest benchmark average among the tested alternatives. The hard-alignment comparison supports matching semantically corresponding content across token positions.}
\label{tab:rebuttal_similarity_alignment}
\begin{tabular}{lccccccc}
\toprule
Metric / alignment & MATH-500 & GSM8K & AIME25 & AMC & Minerva & OlympiadBench & Avg. \\
\midrule
Cosine + soft alignment (default) & \textbf{65.6} & 76.3 & \textbf{3.3} & \textbf{34.9} & 29.8 & 22.1 & \textbf{38.68} \\
Jaccard + soft alignment & 63.4 & 77.6 & 0.0 & 31.3 & 29.4 & \textbf{23.3} & 37.50 \\
Negative $L_2$ + soft alignment & 64.8 & 75.4 & \textbf{3.3} & 30.1 & \textbf{30.1} & 22.1 & 37.63 \\
Inner product + soft alignment & 64.0 & 75.8 & \textbf{3.3} & \textbf{34.9} & 29.0 & 21.7 & 38.12 \\
Cosine + hard alignment & 62.6 & \textbf{78.7} & 0.0 & 33.7 & 28.7 & 20.8 & 37.41 \\
\bottomrule
\end{tabular}%
\end{table}

Table~\ref{tab:rebuttal_similarity_alignment} supports both design choices. \textbf{Similarity metrics:} Cosine similarity, the default metric for \modelname{}, achieves the highest average accuracy among the tested metrics, supporting its use as our default. \textbf{Alignment rules:} Replacing soft alignment with hard alignment lowers the average from 38.68\% to 37.41\%. This supports the use of our soft alignment rule, i.e., matching tokens by semantic similarity across positions: corresponding reasoning content need not occur at the same token position in different rollouts.

\subsection{Sensitivity to the SAE Weight and Alignment Temperature}
\label{app:alpha_tau_sensitivity}

We run additional experiments to analyze the sensitivity of the SAE weight $\alpha$ and alignment temperature $\tau$. Tables~\ref{tab:rebuttal_alpha} and~\ref{tab:rebuttal_tau} show the results. We vary the two hyperparameters independently: Table~\ref{tab:rebuttal_alpha} varies $\alpha$ with $\tau=0.05$, while Table~\ref{tab:rebuttal_tau} varies $\tau$ with $\alpha=2$.

\begin{table}[tbp]
\centering
\small
\setlength{\tabcolsep}{4pt}
\caption{Sensitivity to the SAE weight $\alpha$ at $\tau=0.05$. The default $\alpha=2$ yields the highest average among the tested weights, supporting a moderate contribution from the semantic signal.}
\label{tab:rebuttal_alpha}
\begin{tabular}{lccccccc}
\toprule
Setting & MATH-500 & GSM8K & AIME25 & AMC & Minerva & OlympiadBench & Avg. \\
\midrule
Default ($\alpha=2$) & 65.6 & \textbf{76.3} & \textbf{3.3} & \textbf{34.9} & 29.8 & 22.1 & \textbf{38.68} \\
$\alpha=0.5$ & 65.4 & 74.6 & \textbf{3.3} & 27.7 & 27.2 & 22.6 & 36.80 \\
$\alpha=1$ & \textbf{66.6} & 75.5 & 0.0 & 32.5 & \textbf{30.1} & \textbf{23.6} & 38.06 \\
$\alpha=4$ & 63.0 & 75.9 & 0.0 & 28.9 & 24.3 & 22.6 & 35.77 \\
\bottomrule
\end{tabular}%
\end{table}

The best average accuracy occurs at $\alpha=2$ among the tested values, while increasing the weight to $\alpha=4$ harms performance, lowering the average from 38.68\% to 35.77\%. Thus, stronger SAE guidance is not always better, which supports a moderate weight for $\alpha$.

\begin{table}[tbp]
\centering
\small
\setlength{\tabcolsep}{4pt}
\caption{Sensitivity to alignment temperature $\tau$ at $\alpha=2$. The intermediate temperatures 0.02 and 0.05 outperform both 0.01 and 0.15; $\tau=0.02$ achieves the highest average in this sweep.}
\label{tab:rebuttal_tau}
\begin{tabular}{lccccccc}
\toprule
Setting & MATH-500 & GSM8K & AIME25 & AMC & Minerva & OlympiadBench & Avg. \\
\midrule
Default ($\tau=0.05$) & \textbf{65.6} & 76.3 & 3.3 & 34.9 & 29.8 & 22.1 & 38.68 \\
$\tau=0.01$ & 63.2 & 76.5 & 3.3 & 26.5 & 28.3 & 22.6 & 36.73 \\
$\tau=0.02$ & 61.8 & \textbf{78.6} & \textbf{6.7} & \textbf{36.1} & 29.0 & \textbf{23.7} & \textbf{39.32} \\
$\tau=0.15$ & 63.8 & 70.4 & 3.3 & 33.7 & \textbf{30.1} & 22.7 & 37.33 \\
\bottomrule
\end{tabular}%
\end{table}

For $\tau$, the best average accuracy occurs at $\tau=0.02$, reaching 39.32\% and exceeding our default $\tau=0.05$ by 0.64 percentage points; both outperform the lower temperature of 0.01 and the higher temperature of 0.15. This indicates that alignment sharpness affects performance. The results favor an intermediate temperature among the tested settings, balancing concentration on close matches with contributions from other relevant tokens.

\subsection{Correctness-Balance Weighting and Score Normalization}
\label{app:balance_normalization}

We further run experiments to examine the sensitivity of \modelname{} to the correctness-balance weight $\rho_q$. Since $S$ is standardized over the batch, we vary the shape of the weighting as $S_{qg,t}=\rho_q^{\gamma}(s^+_{qg,t}-s^-_{qg,t})$ for scored tokens. Here $\gamma=0$ removes the weighting, $\gamma=1$ is the default, and $\gamma=2$ strengthens the downweighting of imbalanced correct/incorrect groups. Unscored tokens retain the original zero-SAE-signal convention. Table~\ref{tab:rebuttal_balance} reports the final RL evaluation results of \texttt{Qwen3-\allowbreak{}1.7B-\allowbreak{}Base} on the math task.

\begin{table}[tbp]
\centering
\small
\setlength{\tabcolsep}{4pt}
\caption{Sensitivity to correctness-balance weighting. The default $\gamma=1$ outperforms both removing the weight and squaring it, supporting the original balance between contributions from different rollout groups.}
\label{tab:rebuttal_balance}
\begin{tabular}{lccccccc}
\toprule
Setting & MATH-500 & GSM8K & AIME25 & AMC & Minerva & OlympiadBench & Avg. \\
\midrule
Default ($\gamma=1$) & \textbf{65.6} & \textbf{76.3} & \textbf{3.3} & \textbf{34.9} & \textbf{29.8} & 22.1 & \textbf{38.68} \\
No balance weight ($\gamma=0$) & 63.0 & 73.2 & \textbf{3.3} & 31.3 & 27.6 & \textbf{23.0} & 36.90 \\
Squared balance weight ($\gamma=2$) & 65.2 & 76.1 & 0.0 & 31.3 & 26.1 & 22.4 & 36.85 \\
\bottomrule
\end{tabular}%
\end{table}

Table~\ref{tab:rebuttal_balance} shows that the default weighting achieves 38.68\% average accuracy, compared with 36.90\% without weighting and 36.85\% with squared weighting. This supports using the balance of correct and incorrect rollouts to regulate the contribution of each group's semantic contrast.

We also test our current batch-level normalization method against normalizing the scored tokens of correct solutions and those of incorrect solutions separately. Table~\ref{tab:rebuttal_normalization} shows the final RL evaluation results of \texttt{Qwen3-\allowbreak{}1.7B-\allowbreak{}Base} on the math task.

\begin{table}[tbp]
\centering
\small
\setlength{\tabcolsep}{4pt}
\caption{Joint versus separate score normalization. Normalizing scored tokens together improves the six-benchmark average over separate normalization of correct and incorrect rollouts, supporting the shared batch-level statistics used by \modelname{}.}
\label{tab:rebuttal_normalization}
\begin{tabular}{lccccccc}
\toprule
Setting & MATH-500 & GSM8K & AIME25 & AMC & Minerva & OlympiadBench & Avg. \\
\midrule
Joint normalization (SAEPO default) & \textbf{65.6} & \textbf{76.3} & \textbf{3.3} & \textbf{34.9} & \textbf{29.8} & 22.1 & \textbf{38.68} \\
Separate correct/incorrect normalization & 64.0 & 69.1 & \textbf{3.3} & 33.7 & 27.2 & \textbf{22.3} & 36.60 \\
\bottomrule
\end{tabular}%
\end{table}

The sensitivity results show that \modelname{} is generally robust to the tested hyperparameters, and our batch-level normalization performs better than normalizing separately. As shown in Table~\ref{tab:rebuttal_normalization}, joint normalization achieves 38.68\% average accuracy, compared with 36.60\% for separate normalization, an improvement of 2.08 percentage points.

\subsection{Results across Random Seeds}
\label{app:multiple_seeds}

Next, we compare the final evaluation results of \modelname{}, vanilla GRPO, and the strongest baseline, THR, using multiple random seeds. We use \texttt{Qwen3-\allowbreak{}1.7B-\allowbreak{}Base} with seeds 42, 43, and 44. Table~\ref{tab:rebuttal_seeds} reports each run's final six-benchmark math average, together with the mean and standard deviation across seeds.

\begin{table}[tbp]
    \centering
    \small
    \setlength{\tabcolsep}{4pt}
    \caption{Final six-benchmark math averages (\%) across three random seeds. \modelname{} outperforms vanilla GRPO and THR at every seed, showing that its accuracy gains are consistent across these runs.}
    \label{tab:rebuttal_seeds}
    \begin{tabular*}{\linewidth}{@{\extracolsep{\fill}}lcccc@{}}
        \toprule
        Method & Seed 42 & Seed 43 & Seed 44 & Mean $\pm$ s.d. \\
        \midrule
        Vanilla GRPO
        & 35.56 & 36.29 & 36.25
        & $36.03 \pm 0.41$ \\
        THR
        & 36.72 & 36.58 & 37.02
        & $36.77 \pm 0.22$ \\
        \textbf{\modelname{}}
        & \textbf{38.68} & \textbf{39.82} & \textbf{39.47}
        & $\mathbf{39.32 \pm 0.58}$ \\
        \bottomrule
    \end{tabular*}%
    
\end{table}

\modelname{} achieves a mean accuracy of $39.32\pm0.58$\%, compared with $36.03\pm0.41$\% for vanilla GRPO and $36.77\pm0.22$\% for THR. It improves the mean by 3.29 and 2.55 percentage points, respectively, and ranks highest at all three seeds. Two-sided paired $t$-tests across seeds yield $p=0.0207$ and $p=0.0014$ for comparisons against THR and vanilla GRPO, respectively, both lower than $0.05$. This indicates that both reported improvements are statistically significant.

\section{SAE Quality}
\label{app:sae_quality}

Table~\ref{tab:sae_quality} reports SAE reconstruction quality on MATH rollouts generated by each backbone. For each completion token, we compute the cosine similarity between its residual-stream activation and its SAE reconstruction, then average over all evaluated tokens. We exclude the prompt using \texttt{prompt\_len} and truncate the completion at \texttt{Problem:} if the marker appears.

For Qwen3-1.7B-Base, Qwen3-4B-Base, and OLMo-3-7B, we collect activations after decoder blocks 21, 27, and 21, respectively. We record activations in \texttt{bf16} and compute cosine similarities in \texttt{float32}. Because BatchTopK selects latent activations across tokens in a batch, we use the same batch sizes as in SAE training: 4,096 tokens for Qwen3-1.7B-Base, 2,048 for Qwen3-4B-Base, and 1,024 for OLMo-3-7B.

\begin{table}[!htbp]
\centering
\caption{SAE reconstruction quality on MATH.}
\label{tab:sae_quality}
\begin{tabular}{lc}
\toprule
Backbone & Reconstruction cosine \\
\midrule
\texttt{Qwen3-\allowbreak{}1.7B-\allowbreak{}Base} & 0.960  \\
\texttt{Qwen3-\allowbreak{}4B-\allowbreak{}Base} & 0.799 \\
\texttt{OLMo-\allowbreak{}3-\allowbreak{}7B} & 0.847\\
\bottomrule
\end{tabular}
\end{table}

\subsection{Cross-Task Transfer of the SAE and Its Stability during RL}
\label{app:sae_transfer_stability}

The SAE for \texttt{Qwen3-\allowbreak{}1.7B-\allowbreak{}Base} is trained on a general corpus (FineWeb-Edu web text), and the same SAE is used without fine-tuning for both math and code, i.e., without any task-specific adaptation. To show how consistent a general-corpus SAE is across tasks, we ran additional tests to measure its reconstruction quality on the base model's own math and code rollouts used for RL training. We report the \emph{reconstruction cosine}, i.e., the cosine similarity between each token's hidden state and its SAE reconstruction. On math and code rollouts, our general-corpus SAE maintains reconstruction cosine similarities of 0.957 and 0.952, respectively. Therefore, a general-corpus SAE provides consistent representations across tasks without task-specific adaptation.

We ran additional experiments to evaluate reconstruction quality before and after RL training. Our results show that the frozen SAE maintains high reconstruction quality and substantial overlap in frequently activated features as the policy evolves during RL.

\textbf{Details on the Experiments.} We feed identical code and math rollouts to \texttt{Qwen3-\allowbreak{}1.7B-\allowbreak{}Base} checkpoints before and after RL, so changes in SAE activations reflect changes in the model rather than the input text. Besides reconstruction cosine similarity, we measure \emph{concept overlap (top-1000)} with the base model (step 0): the fraction of its 1,000 most frequently activated SAE features that remain among the 1,000 most frequent after RL. Table~\ref{tab:rebuttal_sae_stability} shows the results.

\begin{table}[tbp]
\centering
\small
\setlength{\tabcolsep}{4pt}
\caption{The frozen SAE maintains high reconstruction cosine and substantial overlap in frequently activated features across tasks and RL checkpoints. Final policies retain 94.3--97.1\% of the base model's top-1,000 active features, supporting reuse of the pretrained SAE throughout policy optimization.}
\label{tab:rebuttal_sae_stability}
\vskip 0.2cm
\begin{tabular}{llcc}
\toprule
Task & Model checkpoint & Reconstruction cosine & Concept overlap (top-1000) \\
\midrule
Math & Base model & 0.957 & 100\% \\
 & SAEPO + GRPO (Final) & 0.932 & 94.3\% \\
 & SAEPO + GSPO (Final) & 0.940 & 94.8\% \\
\midrule
Code & Base model & 0.952 & 100\% \\
 & SAEPO + GRPO (Final) & 0.945 & 96.6\% \\
 & SAEPO + GSPO (Final) & 0.949 & 97.1\% \\
\bottomrule
\end{tabular}%
\end{table}

\textbf{Results.} From Table~\ref{tab:rebuttal_sae_stability}, we can see that reconstruction quality remains high after training: cosine similarity decreases modestly from 0.957 to 0.932--0.940 for math and from 0.952 to 0.945--0.949 for code. Final checkpoints also retain 94.3--97.1\% of the base model's top-1,000 features, indicating that the most frequently activated concepts remain largely consistent. Together, these results support the robustness of the pretrained, frozen SAE to policy evolution during RL.

\section{Case Study Examples}\label{appsec:case_study}

\begin{figure}[H]
  \centering
  \includegraphics[width=\linewidth,height=0.85\textheight,keepaspectratio]
  {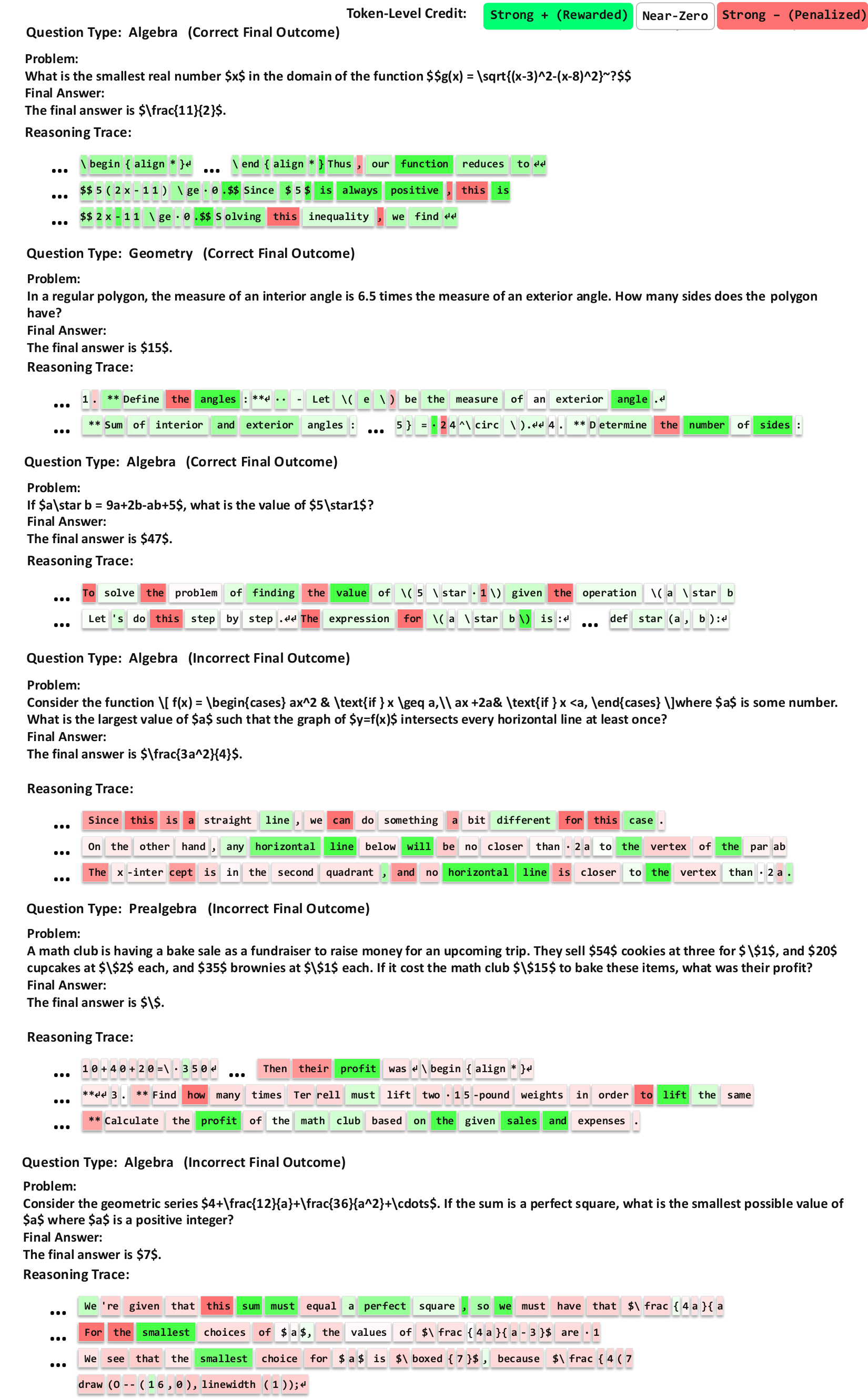}

   \caption{
    More Case Study Examples.
}
   \label{fig:appendix}

\end{figure}

\section{More Top Feature Analysis}

\begin{figure}[H]
  \centering
  \includegraphics[width=0.90\linewidth]
  {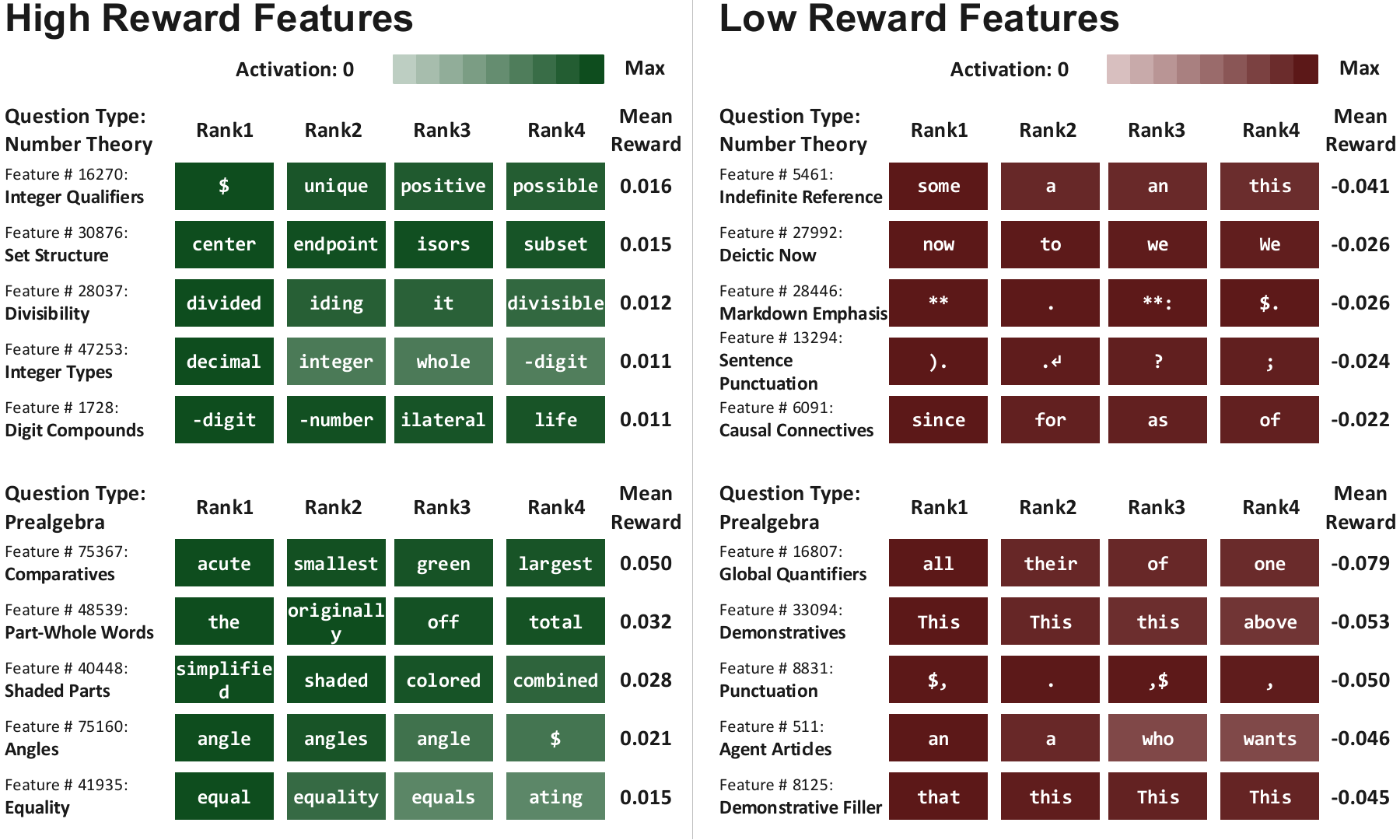}

   \caption{
    More feature analysis results on ``Number Theory'' and ``Prealgebra'' subjects in MATH-500.
}
   \label{fig:app_heatmap1}

\end{figure}

\begin{figure}[H]
  \centering
  \includegraphics[width=0.90\linewidth]
  {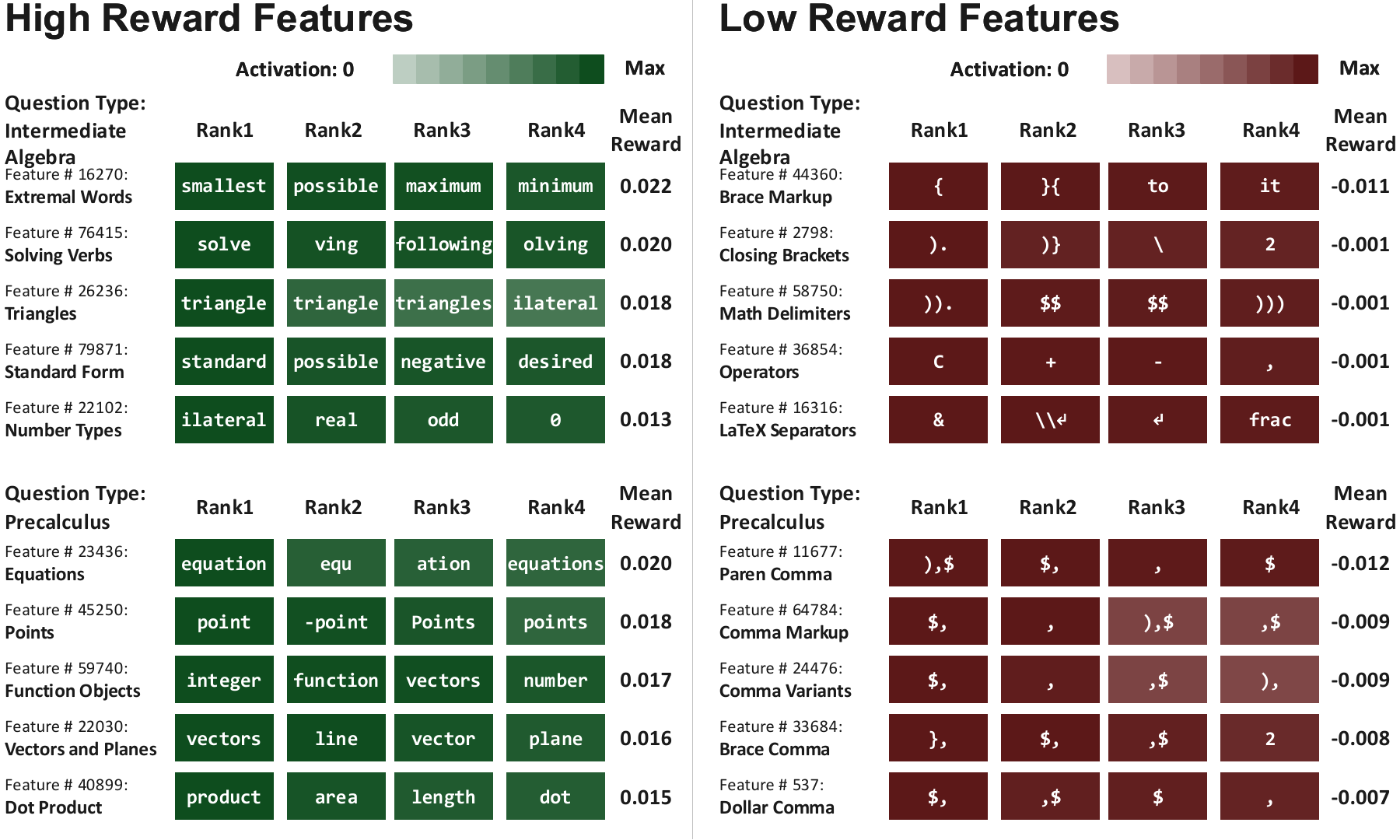}

   \caption{
    More feature analysis results on ``Intermediate Algebra'' and ``Precalculus'' subjects in MATH-500.
}
   \label{fig:app_heatmap2}

\end{figure}

\end{document}